\documentclass{article} 
\usepackage{iclr2027_conference,times}

\usepackage{amsmath,amsfonts,bm}

\def\eqref#1{equation~\ref{#1}}

\def\1{\bm{1}}

\DeclareMathAlphabet{\mathsfit}{\encodingdefault}{\sfdefault}{m}{sl}
\SetMathAlphabet{\mathsfit}{bold}{\encodingdefault}{\sfdefault}{bx}{n}

\usepackage[utf8]{inputenc}
\usepackage[T1]{fontenc}
\usepackage{hyperref}
\usepackage{url}
\usepackage{booktabs}
\usepackage{amsfonts}
\usepackage{nicefrac}
\usepackage{microtype}
\usepackage{xcolor}
\usepackage{algorithm}
\usepackage{algpseudocode}
\usepackage{amssymb}
\usepackage{amsmath}
\usepackage{amsthm}
\newtheorem{lemma}{Lemma}

\newtheorem{proposition}{Proposition}
\newtheorem{remark}{Remark}
\usepackage{graphicx}
\usepackage{multirow}

\graphicspath{{figures/}}

\usepackage{standalone}
\usepackage{subcaption}
\usepackage{float}
\usepackage{tikz}
\usetikzlibrary{positioning, arrows.meta, shapes.geometric, backgrounds, fit}
\usepackage{changepage}
\usepackage{pgfplots}
\pgfplotsset{compat=1.18}
\usepgfplotslibrary{statistics}

\title{%
\centering
\fontsize{15.9}{18}\selectfont
Beyond Gaussian Assumptions: Distribution-Aware\\
Channel Capacity for Effective Connectivity
}

\author{%
\parbox[t]{\dimexpr\textwidth-2\tabcolsep\relax}{%
\centering
\normalfont
\textbf{%
Jianan~Jian\textsuperscript{1},
Jacob~Kang\textsuperscript{1},
Nurahmed~Multezem\textsuperscript{3},
Benjamin~Li\textsuperscript{2},
Nan~Xu\textsuperscript{1,3,4,5}\thanks{Corresponding author: Nan Xu (\texttt{nanxu@umd.edu}).
}%
}\\[6pt]
\normalsize
\mbox{\textsuperscript{1}Fischell Department of Bioengineering}\quad
\mbox{\textsuperscript{2}Department of Computer Science}\\[2pt]
\mbox{\textsuperscript{3}Department of Electrical and Computer Engineering}\quad
\mbox{\textsuperscript{4}Brain and Behavior Institute}\\[2pt]
\mbox{\textsuperscript{5}Artificial Intelligence Interdisciplinary Institute at Maryland}\\[2pt]
University of Maryland, College Park, MD, USA
}%
}

\iclrfinalcopy

\begin{document}

\addtocontents{toc}{\protect\setcounter{tocdepth}{-1}}

\maketitle
\lhead{}
\begin{abstract}
Effective-connectivity estimation from brain signals often relies on Gaussian residual modeling, which enables tractable estimation but can discard informative distributional structure and distort inferred directed interactions when empirical residuals are non-Gaussian. We show across multiple modalities, species, and experimental conditions that both brain signals and fitted channel residuals frequently deviate from Gaussianity. We therefore introduce a distribution-aware, information-theoretic measure of effective connectivity based on channel capacity under general residual distributions. To estimate the resulting capacity from empirical, potentially non-Gaussian residuals, we develop a dual-flow min–max estimator based on normalizing flows, in which a generator searches over admissible input distributions under a power constraint while an observer estimates output entropy. We provide a theoretical characterization of the estimator, showing that the observer objective recovers differential entropy up to a KL approximation term, that the formulation reduces to classical Gaussian capacity as a special case, and that residual entropy can alter achievable information rates beyond variance; game-gap and error analyses further characterize optimization and approximation sources. In brain-like simulations with known directed connectivity, Dual-flow achieves the highest AUROC and AUPRC across ten conditions spanning diverse network topologies, hidden drivers, feedback, and heterogeneous hemodynamics, compared with Gaussian capacity, Granger causality, VAR-LiNGAM, and GIMME. Applied to multimodal brain signals, the method reveals time- and condition-resolved directed interactions consistent with known neurobiological circuitry. Together, these results establish a principled distribution-aware framework for effective-connectivity estimation beyond Gaussian residual modeling.\footnote{Code is available at: https://github.com/inspirelab-site/EC-dual-flow}
\end{abstract}

\section{Introduction}
Estimating time-resolved directed interactions among brain regions is a central problem in computational neuroscience and neuroimaging \citep{friston2011functional}. Effective-connectivity (EC) methods aim to move beyond undirected association by quantifying how activity in one region predicts, drives, or transmits information to another \citep{friston2011functional,valdes2011effective,bielczyk2019disentangling}. Such directed measures are particularly important for understanding dynamic brain states, task-evoked processing, and multimodal brain-signal measurements, including BOLD fMRI, EEG, LFP, and optical neuroimaging \citep{seth2015granger,pan2011broadband,lake2020simultaneous,vafaii2024multimodal}.

{EC estimation is model-based: different approaches define method-specific measures of directed interaction under different assumptions~\citep{valdes2011effective,deshpande2012investigating,bielczyk2019disentangling}.} A common strategy for making EC estimation tractable is to impose Gaussian, linear, or second-order statistical assumptions \citep{granger1969investigating,lutkepohl2005new,barnett2014mvgc,friston2003dynamic,stephan2010ten}. These assumptions underlie many practical estimators and enable closed-form analysis or stable finite-sample estimation \citep{barnett2014mvgc,friston2003dynamic,cover2006elements}. However, whether such Gaussian assumptions are empirically justified across diverse brain-signal modalities remains an open question. Brain signals may exhibit heavy-tailed, skewed, burst-like, or state-dependent structure arising from latent brain states, physiological variability, nonlinear measurement processes, or artifacts \citep{zhang2020relationship,mitra2018spontaneous,mandino2025multimodal,vafaii2024multimodal}. If such distributional structure is present, Gaussian estimators may obscure or distort directed interactions \citep{seth2015granger,friston2003dynamic,stephan2010ten}. Prior work has often adopted modality-specific distributional assumptions for tractable analysis---for example, Rician models for MRI magnitude noise \citep{gudbjartsson1995rician,noh2011rician,wegmann2017bayesian} and approximate Gaussian models in many resting-state fMRI and electrophysiology analyses \citep{granger1969investigating,lutkepohl2005new,barnett2014mvgc,pan2011broadband}. Non-Gaussian approaches such as LiNGAM exploit non-Gaussian disturbances to identify causal structure~\citep{shimizu2006linear,hyvarinen2013pairwise}. However, a systematic cross-modal diagnosis of brain-signal and fitted channel-residual distributions, especially in the context of EC estimation, has received relatively little attention.

\begin{figure}[t]
    \centering
    \includegraphics[width=\linewidth]{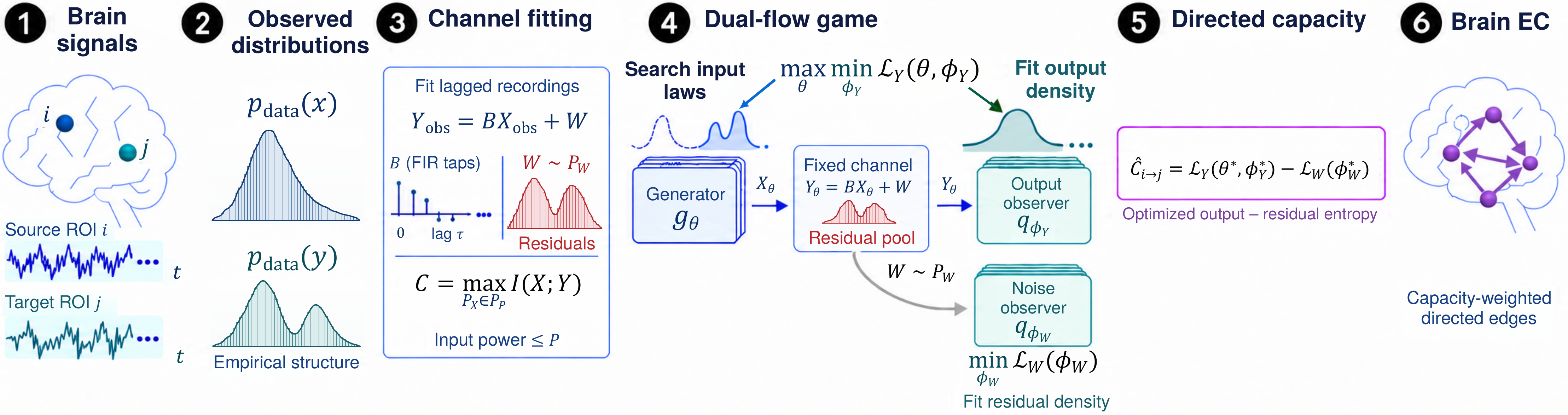}\vspace{-2mm}
    \caption{\textbf{Proposed distribution-aware channel-capacity framework for EC.} Steps 1 \& 2: Select the source and target ROIs and retrieve their signals. Step 3: Perform a linear regression to obtain the linear channel model and a residual pool. Steps 4 \& 5: Compute the channel capacity using a dual-flow minimax game. Step 6: Repeat for all directed ROI pairs to obtain the EC graph.}
    \label{fig:overview}
\end{figure}
Channel capacity quantifies the maximum information rate supported by a channel under specified input constraints \citep{shannon1948mathematical,cover2006elements,gallager1968information}. In EC analysis, a directional interaction between brain regions can be modeled as an empirical communication channel, whose capacity provides an information-theoretic measure reflecting both channel dynamics and the residual distribution \citep{cover2006elements,jian2026channel}. This allows EC estimates to capture distributional structure beyond second-order statistics. For linear additive Gaussian channels, capacity can be computed analytically or semi-analytically \citep{cover2006elements,gallager1968information}. However, Gaussian formulas characterize residual uncertainty through covariance alone and may miss capacity differences arising from non-Gaussian residual structure.

Estimating channel capacity beyond Gaussian settings is challenging because it requires both optimizing over admissible input distributions and estimating output entropy under unknown, potentially non-Gaussian noise \citep{cover2006elements,gallager1968information}. Classical nonparametric entropy estimators can be unstable in finite samples and are difficult to embed inside this optimization \citep{kozachenko1987sample}, while neural mutual-information estimators typically target a fixed joint distribution rather than the capacity-achieving input distribution \citep{belghazi2018mine,poole2019variational}. Thus, a practical estimator must support flexible distribution modeling while remaining compatible with constrained capacity maximization \citep{rezende2015variational,dinh2016density}.

In this work, we challenge the Gaussian assumption in brain-signal EC analysis. We assess brain-signal Gaussianity across multiple modalities and species and examine the distributions of fitted channel residuals. Motivated by the observed deviations, we introduce a distribution-aware, information-theoretic EC measure based on channel capacity. We model each directed pair of regions of interest (ROIs) as a finite impulse response (FIR) channel, using fitted residuals as samples from an unknown continuous noise distribution. To estimate this measure, we develop a dual-flow min--max formulation that couples power-constrained input-distribution optimization with output-entropy estimation. The formulation recovers Gaussian capacity as a special case. Its assumptions of local linear time-invariance and additive white noise may limit applicability to strongly nonlinear dynamics or temporally correlated residuals. This proposed method is summarized in Fig. \ref{fig:overview}. 

\textbf{Contributions.} We make four contributions. First, we assess Gaussianity across brain-signal modalities and conditions, documenting departures in observed signals and fitted channel residuals. Second, we formulate distribution-aware channel-capacity EC using empirical residual resampling under an unknown continuous noise distribution. Third, we develop a dual-flow adversarial normalizing-flow estimator for capacity beyond Gaussian settings. Fourth, we demonstrate superior directed-edge recovery on simulated brain-like signals with ground-truth EC spanning diverse network topologies and hemodynamic confounds. Applications to brain signals reveal time- and condition-resolved directed interactions consistent with known neurobiological circuitry.

\section{Background and related work}

\paragraph{{EC methods under Gaussian and non-Gaussian assumptions.}}
Widely used EC methods include multivariate autoregressive and Granger-causal models \citep{granger1969investigating,lutkepohl2005new,barnett2014mvgc}, as well as information-theoretic measures such as transfer entropy and directed information \citep{schreiber2000measuring,massey1990causality}. Many tractable estimators in these families rely on Gaussian residual assumptions or second-order statistics, which may inadequately capture heavy-tailed, skewed, or state-dependent distributions in empirical brain signals and fitted residuals. Although modality-specific distributional models have been studied, systematic cross-modal assessment of these assumptions in EC estimation remains limited. 

Non-Gaussian EC methods also exist. LiNGAM-family methods exploit non-Gaussian disturbances to identify causal directions under structural and independence assumptions~\citep{shimizu2006linear,hyvarinen2013pairwise}. Our framework instead uses the empirical residual distribution to estimate channel-capacity EC while optimizing the input distribution under a power constraint. Gaussian capacity (GCap) assumes Gaussian residuals, whereas our Dual-flow estimator uses empirical residual samples without this assumption. Further background and interpretation of channel-capacity EC are provided in Appendix~\ref{app:capacity}, with the evaluated
baseline methods detailed in Appendix~\ref{app:baseline}.

\paragraph{Normalizing flows for distribution-aware entropy estimation.}
Normalizing flows are invertible density models that transform a simple base distribution into a flexible target distribution while retaining exact likelihood evaluation through the change-of-variables formula \citep{rezende2015variational,dinh2016density}. Because differential entropy can be estimated from log densities, flows provide a practical tool for entropy estimation when the target distribution is unknown and potentially non-Gaussian. In the present work, we build on this property with an adversarial generator--observer flow architecture: the generator searches over admissible input distributions under a power constraint, while the observer estimates output entropy from samples generated by the empirical channel. Unlike Gaussian-capacity estimators, our framework uses empirically resampled residual noise, allowing channel-capacity estimates to reflect non-Gaussian residual structure in brain-signal measurements.

\section{Distribution-aware channel capacity for EC}

\subsection{Empirical FIR Channels with Residual Resampling}
For each temporal window or condition segment, let $\{x_t\}_{t=1}^{T}$ and $\{y_t\}_{t=1}^{T}$ denote brain-signal time series from a source and target ROI, and approximate the directed interaction $x\to y$ by a locally stationary finite impulse response channel capturing finite-lag effects, \vspace{-1mm}
\begin{align}
    y_t = \sum_{\ell=0}^{k-1} b_\ell x_{t-\ell}+w_t,
    \qquad t=k,\ldots,T,
    \label{eq:channel}
\end{align}\vspace{-1mm}
where $b=(b_0,\ldots,b_{k-1})$ is the channel filter, $w_t$ is additive residual noise, and $T$ denotes the number of time points in the analyzed segment. For each ROI pair and segment, we estimate $\widehat b$ by least squares and compute residuals \vspace{-2mm}
\begin{align}
    \widehat w_t
    =
    y_t-\sum_{\ell=0}^{k-1}\widehat b_\ell x_{t-\ell},
    \qquad t=k,\ldots,T.
     \label{eq:channel_residual}
\end{align} 
Unlike Gaussian-noise channel models, we do not impose a parametric form on the residual distribution. We assume that the additive noise $W$ is drawn from an unknown continuous distribution $P_W$ with finite differential entropy, and use the fitted residuals as empirical samples from this distribution. During training, expectations over $P_W$ are approximated by uniformly resampling from the residual pool $\{\widehat w_t\}_{t=k}^{T}$. The residual pool is not treated as a discrete noise distribution; rather, the residuals are finite samples from an unknown continuous residual-noise distribution $P_W$. Thus the differential-entropy objective is defined with respect to the underlying continuous distribution, while empirical residual resampling provides Monte Carlo minibatches for the flow-based estimator. The fidelity of this approximation depends on residual-pool size and representativeness, the adequacy of the fitted FIR model, and approximate within-segment stationarity  (Appendix~\ref{app:algorithm}).

Stacking Eq.~\ref{eq:channel} over valid samples gives the vector channel
\begin{align}
    Y=BX+W,
    \qquad W\sim P_W,
    \label{eq:vector_channel}
\end{align}

where $B$ is the convolution operator induced by the FIR filter. Within each segment, we assume that the additive FIR approximation is locally valid and that $W$ is independent of admissible inputs $X$. Under an average input-power budget $P>0$, the admissible distributions form the set $\mathcal{P}_P=\{P_X:\mathbb{E}_{P_X}\|X\|_2^2\le dP\}$, where $d$ is the input dimension and $P$ is the allowed power per sample. Under this empirical channel formulation,
\begin{align}
    C
    =
    \sup_{P_X\in\mathcal P_P} I(X;Y)
    =
    \sup_{P_X\in\mathcal P_P} h(BX+W)-h(W),
    \qquad W\sim P_W.
    \label{eq:additive_capacity_identity}
\end{align}

Thus capacity estimation reduces to maximizing output entropy while subtracting residual entropy; a derivation of Eq.~\ref{eq:additive_capacity_identity} is given in Appendix~\ref{app:additive_capacity_identity}. This identity also clarifies why residual shape matters: Gaussian capacity depends only on second-order residual statistics, whereas the general additive-channel formulation depends on the full residual entropy.

\subsection{Dual-Flow Capacity Estimator}
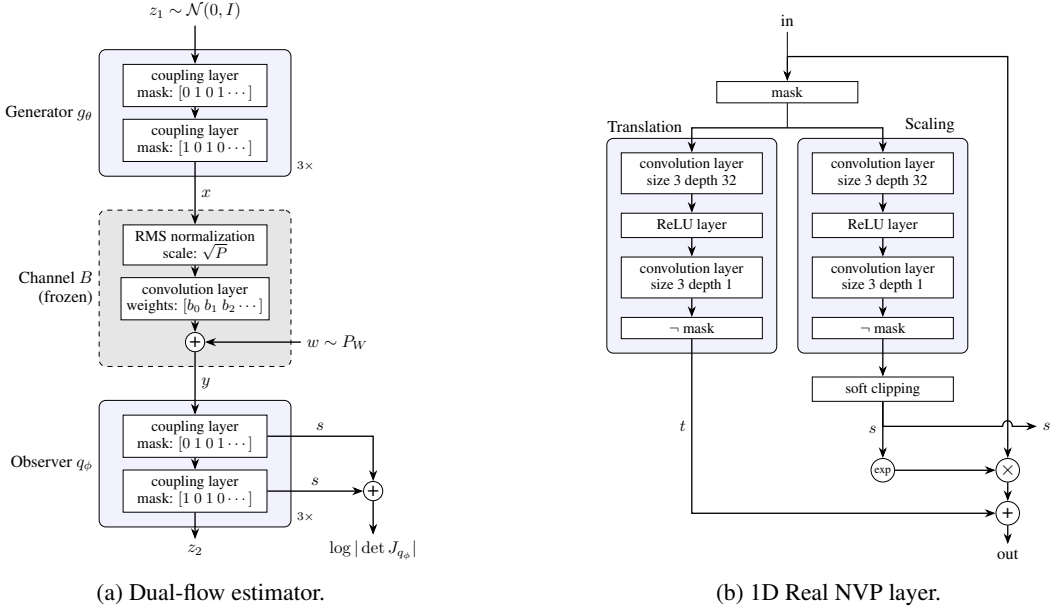
\begin{figure}[h!]
    \centering
    \begin{subfigure}{0.4\linewidth}
        \centering
        \begin{tikzpicture}[
    node distance=2.5mm and 20mm,
    block/.style={rectangle, draw, fill=white, minimum width=30mm, align=center, font=\small},
    frozen/.style={rectangle, draw, fill=gray!20, minimum width=40mm, inner sep=3mm, rounded corners, dashed},
    container/.style={rectangle, draw, fill=blue!5, minimum width=40mm, inner sep=3mm, rounded corners},
    arrow/.style={-Stealth, thick},
    sum/.style={draw, circle, fill=white, inner sep=0pt, minimum size=4mm, font=\large}
]

    \node (z1) {$z_1 \sim \mathcal{N}(0, I)$};
    
    \node (g1) [block, below=8mm of z1] {coupling layer \\ mask: $[0\ 1\ 0\ 1 \cdots]$};
    \node (g2) [block, below=of g1] {coupling layer \\ mask: $[1\ 0\ 1\ 0 \cdots]$};
    \begin{scope}[on background layer]
        \node (gen_box) [container, fit=(g1) (g2), label=left:{Generator $g_\theta$}] {};
        \node [anchor=south west, font=\scriptsize] at (gen_box.south east) {$3\times$};
    \end{scope}
    
    \node (h0) [block, below=10mm of gen_box] {RMS normalization \\ scale: $\sqrt{P}$};
    \node (h1) [block, below=of h0] {convolution layer \\ weights: $[b_0\ b_1\ b_2 \cdots]$};
    \node (a1) [sum, below=of h1] {+};
    \begin{scope}[on background layer]
        \node (chn_box) [frozen, fit=(h0)(a1), label={[align=right]left:{Channel $B$ \\ (frozen)}}] {};
    \end{scope}
    \node (w) [right=of a1]{$w\sim P_W$};
   
    \node (o1) [block, below=10mm of chn_box] {coupling layer \\ mask: $[0\ 1\ 0\ 1 \cdots]$};
    \node (o2) [block, below=of o1] {coupling layer \\ mask: $[1\ 0\ 1\ 0 \cdots]$};
    \begin{scope}[on background layer]
        \node (obs_box) [container, fit=(o1) (o2), label=left:{Observer $q_\phi$}] {};
        \node [anchor=south west, font=\scriptsize] at (obs_box.south east) {$3\times$};
    \end{scope}

    \node (z2) [below=of obs_box] {$z_2$};
    \node (s2) [sum, right=of o2]{+};
    \node (a2) at (z2 -| s2){$\log|\det J_{q_\phi}|$};
    
    \draw [arrow] (z1) -- (g1);
    \draw [arrow] (g1) -- (g2);
    \draw [arrow] (g2) -- node[right] {$x$} (h0);
    \draw [arrow] (h0) -- (h1);
    \draw [arrow] (h1) -- (a1);
    \draw [arrow] (w) -- (a1);
    \draw [arrow] (a1) -- node[right] {$y$} (o1);
    \draw [arrow] (o1) -- (o2);
    \draw [arrow] (o2) -- (z2);
    \draw [arrow] (o1) -| node[above, pos=.25] {$s$} (s2);
    \draw [arrow] (o2) -- node[above, pos=.5] {$s$} (s2);
    \draw [arrow] (s2) -- (a2);

\end{tikzpicture}
        \caption{Dual-flow estimator.}
        \label{fig:fig1a}
    \end{subfigure}
    \hfill
    \begin{subfigure}{0.43\linewidth}
        \centering
        \begin{tikzpicture}[
    node distance=4mm and 4mm,
    block/.style={rectangle, draw, fill=white, minimum width=30mm, align=center, font=\small},
    container/.style={rectangle, draw, fill=blue!5, minimum width=35mm, inner sep=3mm, rounded corners},
    arrow/.style={-Stealth, thick},
    sum/.style={draw, circle, fill=white, inner sep=0pt, minimum size=5mm, font=\large}
]

    \node (in) {in};
    \node (z0) [below=of in] {};
    \node (zm) [block, below=of z0] {mask};
    \node (z1) [below=of zm] {};
    \node (t1) [block, below left=of z1] {convolution layer \\ size 3 depth 32};
    \node (t2) [block, below=of t1] {ReLU layer};
    \node (t3) [block, below=of t2] {convolution layer \\ size 3 depth 1};
    \node (t4) [block, below=of t3] {$\neg$ mask};
    \node (s1) [block, below right=of z1] {convolution layer \\ size 3 depth 32};
    \node (s2) [block, below=of s1] {ReLU layer};
    \node (s3) [block, below=of s2] {convolution layer \\ size 3 depth 1};
    \node (s4) [block, below=of s3] {$\neg$ mask};
    \node (s4c) [block, below=8mm of s4] {soft clipping};
    \node (e1) [sum, below=12mm of s4c, font=\scriptsize] {\strut exp};
    \node (m1) [sum, right=24mm of e1.center] {$\times$};
    \node (a1) [sum, below=of m1] {+};
    \node (z2) [below=of a1] {out};
    
    \begin{scope}[on background layer]
        \node (s_box) [container, fit=(s1) (s4), label={[xshift=10mm]above:{Scaling}}] {}; 
        \node (t_box) [container, fit=(t1) (t4), label={[xshift=-10mm]above:{Translation}}] {};
    \end{scope}
    
    \draw (in) -- (z0.center);
    \draw [arrow] (z0.center) -- (zm) ;
    \draw (zm) -- (z1.center);
    \draw [arrow] (z1.center) -| (t1);
    \draw [arrow] (z1.center) -| (s1);
    \draw [arrow] (t1) -- (t2);
    \draw [arrow] (t2) -- (t3);
    \draw [arrow] (t3) -- (t4);
    \draw [arrow] (s1) -- (s2);
    \draw [arrow] (s2) -- (s3);
    \draw [arrow] (s3) -- (s4);
    \draw [arrow] (s4) -- (s4c);
    \draw [arrow] (s4c) -- node[left] {$s$} (e1);
    \draw [arrow] (e1) -- (m1);
    \draw [arrow] (z0.center) -| (m1);
    \draw [arrow] (m1) -- (a1);
    \draw [arrow] (t4) |- node[left, pos=.25] {$t$} (a1);
    \draw [arrow] (a1) -- (z2);

    \node (s5) [below=of s4c] {};
    \node (s6) [right=24.25mm of s5.center] {};
    \node (s7) [right=.8mm of s6.center] {};
    \node (s8) [right=of s7] {$s$};
    \draw[arrow] (s4c) |- (s6.center) 
          to[out=90, in=90, looseness=2] (s7.center) 
          -- (s8);

\end{tikzpicture}
        \caption{1D Real NVP layer.}
        \label{fig:fig1b}
    \end{subfigure}
    \caption{\textbf{Dual-flow estimator and flow architecture.}
(a) The generator searches over admissible input distributions under a power constraint, while the observer estimates channel-output entropy.
(b) A 1D Real NVP coupling layer used in the generator and observer.}
    \label{fig:fig1}
\end{figure}

Eq.~\ref{eq:additive_capacity_identity} requires optimizing over admissible input distributions while estimating differential entropy under unknown residual noise. We introduce a Dual-flow estimator to address this by combining three complementary ingredients: an information-theoretic channel-capacity objective, a game-theoretic generator--observer interaction, and normalizing flows for flexible non-Gaussian distribution modeling. The generator searches over
power-constrained inputs, while the observer estimates the induced output entropy (Fig.~\ref{fig:fig1}).

The generator $g_\theta$ maps Gaussian latent samples $Z_1\sim\mathcal N(0,I)$ to candidate inputs and enforces the power constraint by RMS normalization,
\begin{align}
    X_\theta=\mathrm{Norm}_P(g_\theta(Z_1)),
    \qquad
    Y_\theta=B X_\theta+W,
    \quad W\sim P_W.
    \label{eq:generated_output}
\end{align}
The observer $q_\phi$ is an invertible flow that induces the density
\begin{align}
    p_\phi(y)
    =
    p_U(q_\phi(y))\left|\det J_{q_\phi}(y)\right|,
    \qquad U\sim\mathcal N(0,I).
    \label{eq:observer_density}
\end{align}
Its negative log-likelihood is
\begin{align}
    L_\phi(y)
    =
    -\log p_\phi(y)
    =
    -\log\left|\det J_{q_\phi}(y)\right|
    +\frac{1}{2}\|q_\phi(y)\|_2^2
    +\frac{d}{2}\log(2\pi).
    \label{eq:observer_loss}
\end{align}

\begin{lemma}[Observer-flow entropy identity]
Let $P_Y$ be the output distribution and let $p_\phi$ be the density induced by the observer flow. Then
\begin{align}
    \mathbb E_{Y\sim P_Y}[-\log p_\phi(Y)]
    =
    h(Y)+\mathrm{KL}(P_Y\|P_\phi).
    \label{eq:entropy_kl_identity}
\end{align}
Consequently, for a sufficiently expressive observer family, minimizing the expected negative log-likelihood estimates $h(Y)$, with equality when $P_\phi=P_Y$.
\end{lemma}

For the signal-plus-noise output and residual-noise terms, define
\begin{align}
    \mathcal L_Y(\theta,\phi_Y)
    &=
    \mathbb E
    \left[
    -\log p_{\phi_Y}
    \left(B\mathrm{Norm}_P(g_\theta(Z_1))+W\right)
    \right], \\
    \mathcal L_W(\phi_W)
    &=
    \mathbb E[-\log p_{\phi_W}(W)].
    \label{eq:flow_losses}
\end{align}
The resulting flow-restricted capacity estimator is
\begin{align}
    \widehat C_{\Theta,\Phi}
    =
    \max_{\theta\in\Theta}
    \min_{\phi_Y\in\Phi}
    \mathcal L_Y(\theta,\phi_Y)
    -
    \min_{\phi_W\in\Phi}
    \mathcal L_W(\phi_W).
    \label{eq:dual_flow_capacity_estimator}
\end{align}
The generator searches for an admissible input distribution that maximizes the observer-estimated output entropy, while the observer estimates the induced output density. We use the adversarial terminology to describe this optimization structure, but do not claim strong duality for the nonconvex neural parameterization.

Both the generator and observer use 1D Real NVP flows with three coupling-layer pairs (Fig.~\ref{fig:fig1}b; architecture details in Appendix~\ref{app:algorithm}). In each training iteration, we generate a training sequence of length \(L\) by sampling a latent sequence \(Z_1\sim\mathcal N(0,I_L)\) and drawing \(L\) residual-noise samples by uniformly resampling from the fitted residual pool. These samples are propagated through the generator, empirical FIR channel, and observer to obtain Monte Carlo estimates of the entropy objectives. We alternate observer updates minimizing negative log-likelihood with generator updates maximizing the signal-plus-noise entropy estimate; gradients are computed by automatic differentiation and parameters are optimized with {\it Adam}. The full training procedure is given in Algorithm~\ref{alg:dual_flow_capacity} (Appendix~\ref{app:algorithm}). Capacity estimates showed limited sensitivity to the generated training-sequence length \(L\) over the tested range (Appendix~\ref{app:L_sensitivity}).


To estimate capacity, we train signal-plus-noise and noise-only objectives in parallel using matched residual samples. The noise-only objective estimates $h(W)$ by applying an observer flow directly to resampled residual noise. This paired estimation reduces Monte Carlo variance in the entropy difference. The final estimate is
\begin{align}
    \widehat C=\widehat h(Y)-\widehat h(W).
\end{align}
Convergence is declared when an exponential moving average of this paired entropy difference reaches stationarity.

\subsection{Key Theoretical Properties}
\label{sec:theory}
We next characterize the key theoretical properties of the Dual-flow formulation. We first establish the observer-flow entropy identity, then show that the formulation recovers classical Gaussian capacity as a special case and that residual distributional structure can affect capacity beyond variance. We further characterize optimization and approximation errors of the generator--observer game. Corresponding controlled checks are reported in Appendix~\ref{app:controlled_validation}.

\paragraph{Entropy estimation by the observer flow.}
Lemma~1 shows that the observer negative log-likelihood equals the target entropy plus a KL approximation error, justifying the minimized observer loss as an entropy estimator inside the capacity objective (Appendix~\ref{app:observer_entropy_identity}). In controlled settings with analytically known noise entropy, the observer accurately recovered the target entropy across Gaussian and non-Gaussian residual laws (Fig.~\ref{fig:fig_sim}A-i; Appendix~\ref{app:controlled_validation}).

\paragraph{Gaussian recovery.}
When $W\sim\mathcal N(0,\Sigma_W)$, the population capacity formulation in Eq.~\ref{eq:additive_capacity_identity} reduces to the classical Gaussian-channel capacity:
\begin{align}
    C_{\mathrm{G}}
    =
    \max_{\Sigma_X\succeq0,\;\mathrm{tr}(\Sigma_X)\le dP}
    \frac{1}{2}
    \log\det
    \left(
        I+
        \Sigma_W^{-1/2}B\Sigma_XB^\top\Sigma_W^{-1/2}
    \right).
    \label{eq:gaussian_capacity_matrix}
\end{align}
Thus the generalized formulation includes classical Gaussian capacity as a special case while allowing the residual uncertainty term $h(W)$ to reflect empirical non-Gaussian structure; the formal derivation is given in Appendix~\ref{app:gaussian_recovery}. In controlled Gaussian FIR channels with known analytic capacity, the full Dual-flow estimator recovered the corresponding capacity across representative channel structures (Fig.~\ref{fig:fig_sim}A-ii; Appendix~\ref{app:controlled_validation}).

\paragraph{Residual entropy beyond residual variance.}
For a scalar additive channel $Y=X+W$ with $\mathbb E[X^2]\le P$, $\mathrm{Var}(W)=\sigma_W^2$, and entropy power $N(W)=(2\pi e)^{-1}\exp(2h(W))$, the entropy power inequality gives the achievable-rate lower bound
\begin{align}
    C(W)
    \ge
    \frac{1}{2}\log\left(1+\frac{P}{N(W)}\right)
    \ge
    \frac{1}{2}\log\left(1+\frac{P}{\sigma_W^2}\right),
    \label{eq:nongaussian_capacity_gap}
\end{align}
with equality in the second inequality only when $W$ is Gaussian. Thus residual distributions with matched variance but different entropy power can support different achievable rates under the same input-power constraint. A matched-Gaussian estimator cannot distinguish these cases because it replaces the residual distribution by a Gaussian approximation with the same second-order statistics. The theoretical result and its FIR extension are derived in
Appendix~\ref{app:entropy_power}; the corresponding controlled matched-variance experiment is shown in Fig.~\ref{fig:fig_sim}B (Appendix~\ref{app:controlled_validation}).

\begin{remark}[Approximation sources]
The finite-sample estimate differs from the population capacity because of residual/channel estimation, generator restriction, observer density approximation, nonconvex optimization, and Monte Carlo sampling. We summarize these sources as
\begin{align}
    |\widehat C_{\Theta,\Phi}-C|
    \lesssim
    \varepsilon_{\mathrm{resid}}
    +\varepsilon_{\mathrm{gen}}
    +\varepsilon_{\mathrm{obs}}
    +\varepsilon_{\mathrm{opt}}
    +\varepsilon_{\mathrm{MC}}.
    \label{eq:error_decomposition}
\end{align}
This decomposition is meant as a practical guide to estimator limitations rather than a finite-sample guarantee. Additional details, including the game-gap diagnostic for the flow-restricted optimization problem and the corresponding error decomposition, are provided in Appendix~\ref{app:game_gap} and Appendix~\ref{app:error_decomposition}.
\end{remark}

\subsection{Time- and Condition-Resolved EC Estimation}
The same empirical residual-aware estimator can be applied either to sliding temporal windows or to pre-defined experimental conditions. For continuously recorded data, such as resting-state LFP--BOLD recordings, we estimate dynamic directed connectivity in sliding windows. For block-structured task data, such as motor fMRI, we estimate condition-resolved directed connectivity across independently defined task and rest windows rather than interpreting the estimates as a continuous temporal trajectory. For each window or condition segment $\tau$ and directed ROI pair $i\to j$, we estimate the FIR filter, resample from the fitted residual pool, and compute
\(\widehat C_{i\rightarrow j}(\tau)\).

The FIR order $k$ is selected separately for each ROI pair and segment. We use BIC for sufficiently long windows and corrected AIC for shorter windows with low sample-to-parameter ratios \citep{hurvich1989regression}. Given the selected $k$, FIR coefficients are estimated by least squares, and the resulting residuals are retained as empirical samples from the unknown residual-noise distribution rather than replaced by a parametric Gaussian model.
\section{Results}
The experiments serve three complementary roles: distributional diagnosis, ground-truth validation, and real-data application. We first diagnose departures from Gaussianity in brain signals across modalities and in fitted rat BOLD channel residuals (Section~\ref{sec:brain-data-distn}; Appendix~\ref{app:gaussianity_analysis}). We then validate directed-edge recovery in brain-like networks with known ground-truth EC across diverse network and hemodynamic conditions (Section~\ref{sec:simulation}; Appendices~\ref{app:simulation_details}). Finally, real-data analyses assess task sensitivity and biological plausibility in tongue-motion fMRI (Section~\ref{sec:tongue-motion}; Appendix~\ref{app:empirical_application}), with complementary neurophysiological evidence from cross-modal correspondence (Appendix~\ref{app:swc}). Computational requirements and runtime are reported in Appendix~\ref{app:cost}.

\subsection{Empirical Brain Signals and Channel Residuals Exhibit non-Gaussian Structure}\label{sec:brain-data-distn}

\begin{figure*}[h!]
\centering
    \begin{subfigure}{0.25\linewidth}
        \centering
        \includegraphics[width=\linewidth]{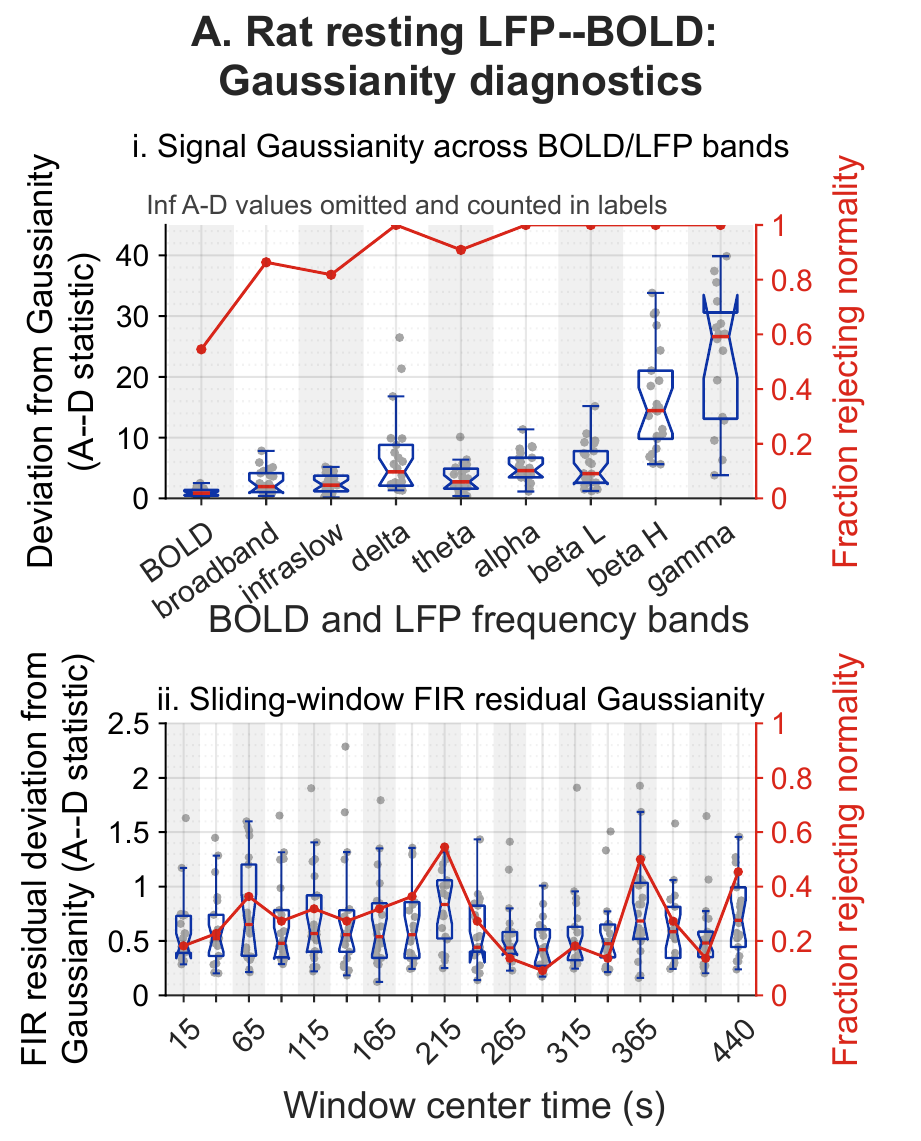} 
    \end{subfigure}
    \begin{subfigure}{0.74\linewidth}
        \centering
        \includegraphics[width=\linewidth]{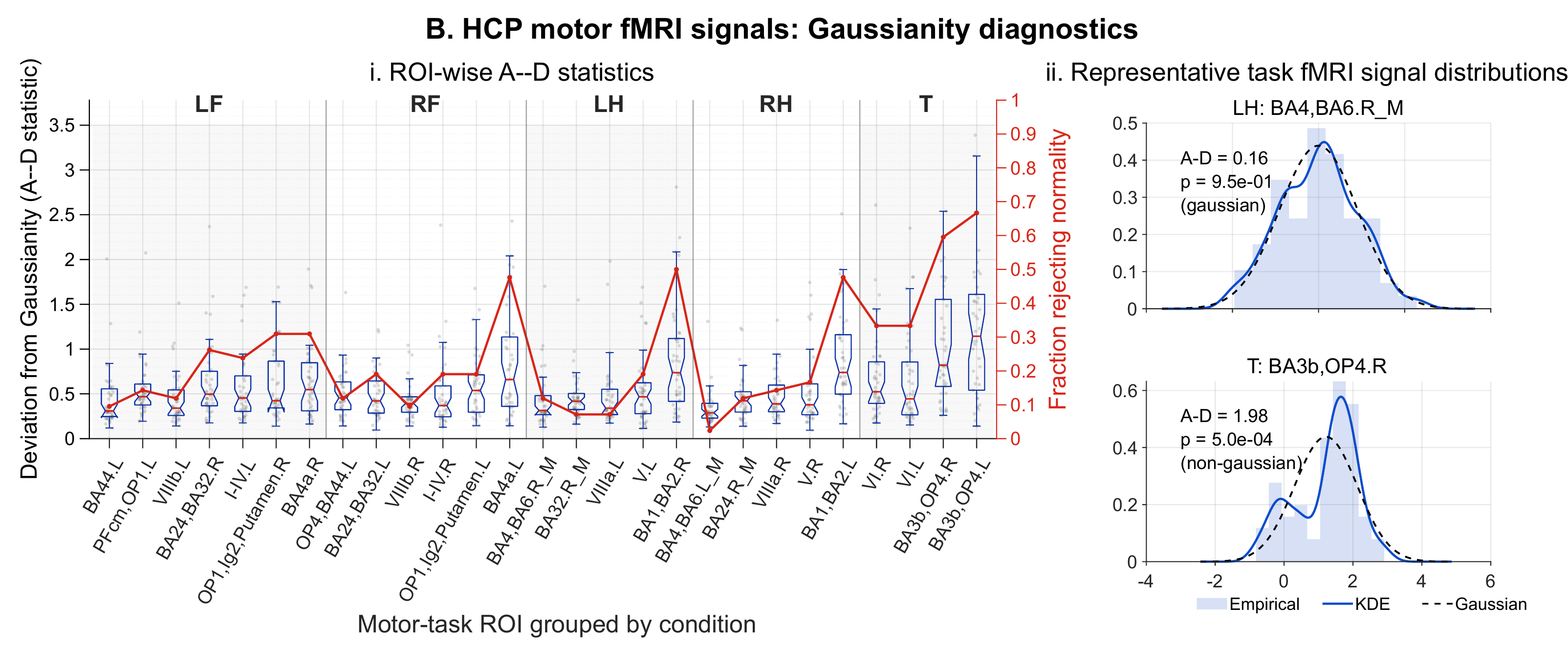} 
    \end{subfigure}
    \begin{subfigure}{\linewidth}
        \centering
        \includegraphics[width=\linewidth]{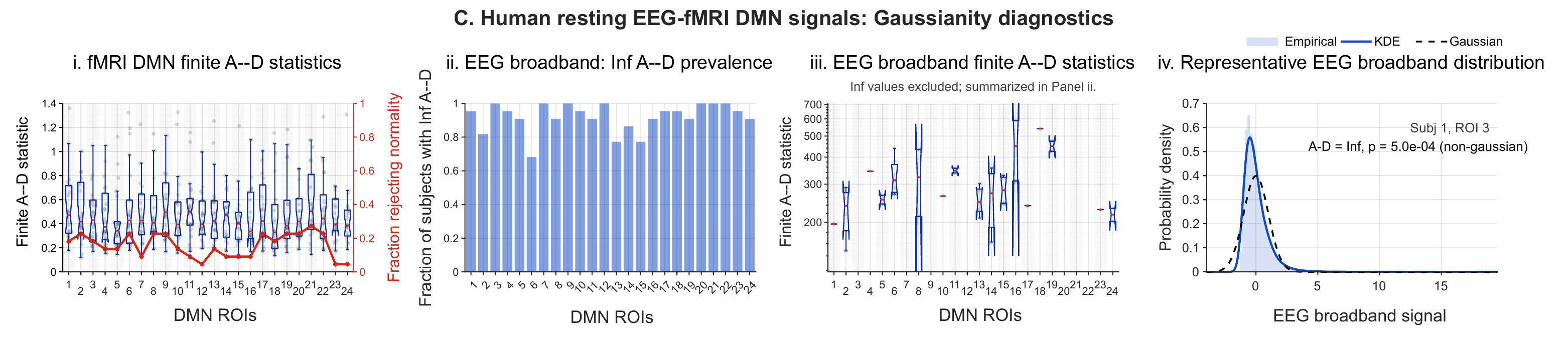}
    \end{subfigure}\vspace{-2mm}
\caption{
\textbf{Gaussianity diagnostics across neural signal modalities and datasets.}
Anderson--Darling (A--D) statistics quantify deviations from matched Gaussian distributions.
\textbf{A}, Rat resting LFP--BOLD signals show frequency-dependent signal non-Gaussianity (\textit{i}); sliding-window FIR residuals show time-varying Gaussianity, with the red curve indicating the fraction of subjects rejecting residual normality (\textit{ii}).
\textbf{B}, HCP motor fMRI shows ROI- and task-dependent Gaussianity variation (\textit{i}) with representative task-ROI distributions and matched Gaussian fits (\textit{ii}).
\textbf{C}, Concurrent EEG--fMRI DMN signals show modest finite A--D statistics for fMRI BOLD (\textit{i}) but widespread infinite A--D statistics for broadband EEG; infinite values are summarized by prevalence (\textit{ii}), excluded from finite-value boxplots (\textit{iii}), and illustrated by a representative distribution (\textit{iv}). Dataset descriptions, preprocessing, and Gaussianity-analysis details are provided in Appendix~\ref{app:gaussianity_analysis}.
\vspace{-.2cm}
}
\label{fig:distn_all}
\end{figure*}

Many EC models assume, either explicitly or implicitly, that model residuals, prediction errors, or observation noise can be adequately characterized by second-order statistics. This assumption is central to linear autoregressive and Granger-causal frameworks, where directed interactions are inferred from reductions in prediction-error variance or covariance under VAR/MVAR models \citep{granger1969investigating,bressler2011wiener,seth2015granger}. Related Gaussian noise assumptions also appear in generative EC frameworks such as dynamic causal modeling, where observation and state noise are commonly modeled within Gaussian state-space or variational inference formulations \citep{friston2003dynamic,stephan2010ten}. 

Across rat resting LFP--BOLD recordings (Fig.~\ref{fig:distn_all}A), HCP motor-task fMRI (Fig.~\ref{fig:distn_all}B), and human resting simultaneous EEG--fMRI datasets (Fig.~\ref{fig:distn_all}C), empirical signal distributions were not uniformly Gaussian. Rat LFP--BOLD recordings showed frequency-dependent deviations from Gaussianity, with stronger departures in higher-frequency LFP bands.  HCP motor fMRI showed ROI- and task-dependent variation in Anderson--Darling (A--D) statistics, and representative task-ROI distributions showed visible mismatch from matched Gaussian fits. In simultaneous EEG--fMRI, BOLD signals from default mode network (DMN) regions of interest (ROIs) showed relatively modest finite A--D statistics, whereas broadband EEG-derived DMN signals frequently produced numerically infinite A--D statistics, indicating severe departures from Gaussianity.

These signal-level diagnostics are directly relevant to EC estimation because Gaussian residual or prediction-error models reduce uncertainty to variance or covariance. Under the additive FIR channel model (Eq.~\ref{eq:vector_channel}), the residual distribution \(P_W\) directly determines the uncertainty term entering capacity estimation. We therefore fit local FIR channels in sliding windows and examined the resulting 100-sample residual time series in each window. In the rat resting BOLD channel, residual A--D statistics varied across windows, and the fraction of subjects rejecting residual normality also changed over time (Fig.~\ref{fig:distn_all}A-ii). Thus, the issue is not only that neural signals can be non-Gaussian, but also that the channel residual distribution can be time-varying and only locally Gaussian-like (see Fig.~\ref{fig:sw_dist} in the Appendix for a sample time-varying noise histogram in the LFP-BOLD dataset). Together, these observations motivate estimating \(P_W\) from empirical residual samples rather than approximating residual uncertainty solely through Gaussian covariance structure. Fig.~\ref{fig:sw_cc} shows that larger signal-level Anderson--Darling statistics were associated with greater differences between GCap and Dual-flow capacity estimates.


\subsection{Brain-Like Simulations with Ground-Truth EC}
\label{sec:simulation}
Because ground-truth EC is unavailable in neuroimaging, we validated Dual-flow's EC-recovery capability across ten brain-like simulation conditions against Gaussian capacity (GCap), its within-framework matched-Gaussian counterpart, as well as GC, VAR-LiNGAM, and GIMME (see Appendix~\ref{app:baseline} for baseline methods). The conditions capture network-level and hemodynamic variation challenges relevant to brain EC estimation, including indirect pathways, hidden drivers, feedback, modular organization, heterogeneous HRFs, their coexistence, and increasing network scale (3--28 ROIs; 50 realizations per condition). Simulation design and ground-truth EC topologies are detailed in Appendix~\ref{app:sim_topologies}, with the evaluation metrics in Appendix~\ref{app:sim_protocol}.

\begin{figure}[h!]
    \centering
    \includegraphics[width=\linewidth]
    {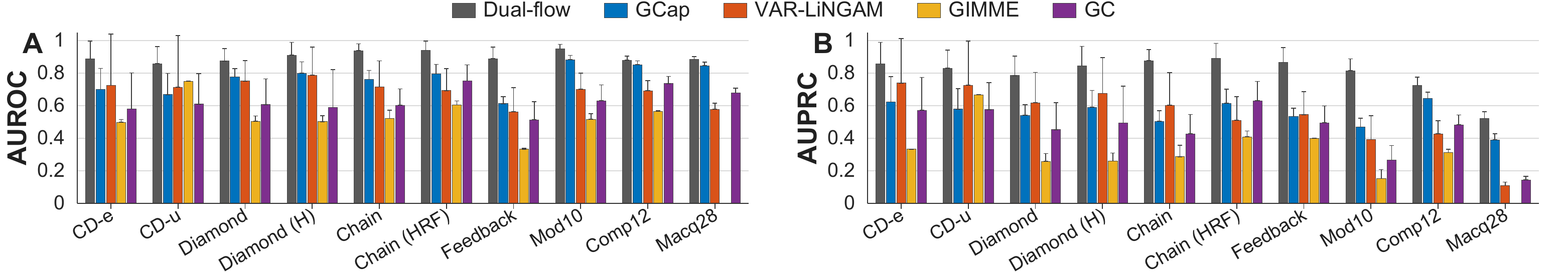}
    \caption{\textbf{Directed-edge recovery across brain-like simulations.} Bars show mean AUROC (\textbf{A}) and AUPRC (\textbf{B}) across 50 realizations per condition; error bars denote $+1$ SD (upper half shown for clarity). Dual-flow achieved the highest mean AUROC and AUPRC in every condition. Full results are reported in Appendix~\ref{app:sim_full}.}
    \label{fig:sim_main}
\end{figure}
Dual-flow achieved the highest AUROC and AUPRC in all ten conditions (AUROC .858--.949; AUPRC .521--.891; Fig.~\ref{fig:sim_main}). In the common-driver conditions, edge ranking remained strong despite larger variability in threshold-dependent metrics; because each realization contains only two true and four null edges, sensitivity and FPR change in coarse increments of .50 and .25, respectively, making these metrics intrinsically discrete. The Diamond, hidden-Diamond, and Chain conditions tested robustness to convergent and indirect pathways and to an unobserved common driver, while Chain (HRF) isolated regional hemodynamic heterogeneity. The Feedback and Mod10 conditions extended the evaluation to reciprocal/cyclic interactions and modular organization. Performance remained strong when these challenges coexisted in Comp12 (AUROC/AUPRC .879/.724) and in the sparse 28-ROI recurrent Macq28 network (.884/.521), where only 52 of 756 candidate directed edges were true (random AUPRC .069). Together, these results show that Dual-flow preserves directed-edge ranking across increasingly challenging network and hemodynamic conditions, including their coexistence and larger network scale. Full results with additional metrics are reported in Appendix~\ref{app:sim_full}.
\subsection{Distribution-Aware Capacity Detects Directed Interactions in Tongue-Motion fMRI}\label{sec:tongue-motion}

We applied the proposed estimator to the tongue-motion condition of the motion-evoked fMRI dataset, as this condition showed the strongest departure from Gaussianity among the movement conditions in Fig.~\ref{fig:distn_all}B-i (see Appendix~\ref{app:empirical_application} for implementation details). For each subject, condition, and directed ROI pair, we fit a local FIR channel and computed four directed-connectivity estimates: Dual-flow, GCap, GC, and VAR-LiNGAM. This analysis evaluates whether Dual-flow can detect task-related directed interactions beyond GCap, GC, and VAR-LiNGAM in a condition whose signals showed relatively strong departure from Gaussianity, rather than focusing on within-run temporal variability. Following established hypothesis-driven EC practice that restricts analysis to task-relevant ROIs and prespecified circuits \citep{smith2012effective,deshpande2012investigating}, we predefined four tongue-motion ROIs from the HCP motor dataset's task-activation maps, following its established activation-mapping framework \citep{barch2013function}; details are provided in Appendix~\ref{app:gaussianity_analysis}.



\begin{figure}[h!]
    \centering
    \includegraphics[width=300pt]{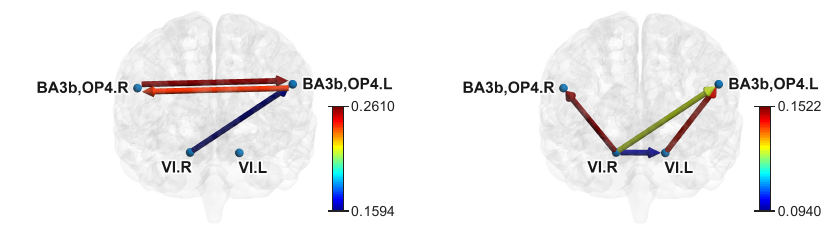}
    \caption{\textbf{Task-related directed connections of Dual-flow (left) and GCap (right)} visualized in BrainNet Viewer \citep{xia2013brainnet}. Result with all four EC metrics, as well as their respective connectivity matrices in the task state and rest states, are reported in Fig.~\ref{fig:brainnet_all}.}
    \label{fig:HCP}
\end{figure}

For the tongue-motion task, prior literature on contralateral cerebello-cortical organization and bilateral orofacial sensorimotor control supports four physiologically plausible candidate interactions: (i) VI.R $\longrightarrow$ BA3b, OP4.L; (ii) VI.L $\longrightarrow$ BA3b, OP4.R; (iii) BA3b, OP4.L $\longrightarrow$ BA3b, OP4.R; and (iv) BA3b, OP4.R $\longrightarrow$ BA3b, OP4.L~\citep{manto2012consensus,sasegbon2023role,rizzolatti2001cortical,todorov2004optimality}. These literature-supported interactions provide a physiologically plausible reference but not ground-truth labels for the exact directed edges in the present fMRI data. Task-sensitive edges that were estimated by each EC method were required to meet both FDR-corrected tongue-versus-rest significance (\(q<0.1\)) and a prespecified cross-subject stability criterion (\(\mathrm{CV}\) below the 30th percentile); see Appendix~\ref{app:empirical_application} for implementation details and Table~\ref{tab:tongue_stats} for numerical results. As shown in Fig.~\ref{fig:HCP}, Dual-flow identified three of the four candidates---(i), (iii), and (iv)---while the missing mirrored cerebello-cortical interaction (ii) may reflect asymmetry in the sampled data or limited statistical power. GCap did not identify interaction (iii) and instead identified connections between the left cerebellum and left sensorimotor cortex; this difference is consistent with the stronger departure from Gaussianity observed in motor-cortical signals (Fig.~\ref{fig:distn_all}B-i), although the real-data analysis does not isolate non-Gaussianity as its sole cause. GC and VAR-LiNGAM did not identify any significant tongue-evoked directed connections (Table.~\ref{tab:tongue_stats}).

\section{Conclusion}
We introduced a distribution-aware channel-capacity framework that incorporates the distributional structure of brain signals and fitted residuals into effective-connectivity (EC) estimation. Rather than reducing residual uncertainty to Gaussian second-order statistics, Dual-flow uses empirical residual samples and optimizes the admissible input distribution to estimate the information-carrying capacity of a fitted directional interaction under general residual distributions. Across modalities and conditions, brain-signal and fitted-residual distributions frequently departed substantially from Gaussianity, providing empirical support for explicitly incorporating distributional structure into EC estimation. We theoretically characterize the framework by establishing Gaussian-capacity recovery, the effect of residual structure beyond variance on achievable information rates, and the main approximation and optimization errors. In brain-like simulations with known ground-truth EC, Dual-flow showed strong directed-edge recovery across diverse network and hemodynamic challenges and consistently outperformed alternative baselines. Empirical applications further demonstrate its potential as a practical tool for time- and condition-resolved EC analysis, with potential value for both basic neuroscience and translational studies of task-evoked or disrupted effective connectivity. \paragraph{Limitations.} The current formulation assumes a linear FIR channel with additive noise within each temporal window or condition segment, which may be inadequate for strongly nonlinear dynamics. The current implementation is designed for hypothesis-driven EC analysis of prespecified circuits and moderate-sized subnetworks, an established use case in cognitive and clinical neuroimaging \citep{smith2012effective,deshpande2012investigating,bielczyk2019disentangling}. A remaining limitation is that it does not yet scale efficiently to unrestricted whole-brain exploratory analysis, which represents a distinct use case aimed at broader network-discovery questions \citep{deshpande2012investigating,bielczyk2019disentangling}.

\subsection*{Acknowledgement}
This research is supported by NIH R00NS123113. Data were provided in part by the Human Connectome Project, WU-Minn Consortium (Principal Investigators: David Van Essen and Kamil Ugurbil; 1U54MH091657) funded by the 16 NIH Institutes and Centers that support the NIH Blueprint for Neuroscience Research; and by the McDonnell Center for Systems Neuroscience at Washington University. We also thank Dr. Keilholz group in providing the LFP-fMRI data in rats. We used generative AI tools to refine Figure 1 and improve the organization and wording of the manuscript. The authors reviewed all AI-assisted revisions and take responsibility for the final content.
\bibliography{iclr2027_conference}
\bibliographystyle{iclr2027_conference}

\appendix
\addtocontents{toc}{\protect\setcounter{tocdepth}{2}}
\clearpage
\begingroup
  \renewcommand{\contentsname}{Appendix Contents}
  \setcounter{tocdepth}{2}
  \tableofcontents
\endgroup
\clearpage

\section{Background and Related Work}
\label{app:background}

\subsection{Channel Capacity for Effective Connectivity: Motivation and Interpretation}
\label{app:capacity}

Effective connectivity (EC) is inherently model-based: different approaches define directed interaction through different models, assumptions, and statistical objectives \citep{roebroeck2005mapping,valdes2011effective,deshpande2012investigating,bielczyk2019disentangling}. For example, GC measures directed predictability, whereas structural equation modeling and LiNGAM-based methods infer directed relationships under different assumptions about network structure and residual distributions. Accordingly, their outputs are not directly interchangeable measures of connection strength. \citet{smith2011network} compared such methods by their ability to recover known edges and directions from simulated fMRI. Following this practice, we introduce channel capacity not as a universal measure of EC strength, but as a distinct information-theoretic measure evaluated through directed-edge recovery.

Brain-signal and residual distributions are not nuisance details; they directly shape the information-carrying limit of a fitted directional interaction. Our framework incorporates this distributional structure directly into a channel-capacity EC measure. For each ordered ROI pair \(X\rightarrow Y\), we fit a directional finite impulse response (FIR) model and interpret its fitted dynamics and residuals as a noisy communication channel. Under an input-power constraint, capacity quantifies the maximum information rate supported by the fitted channel, reflecting its directional dynamics and residual distribution while optimizing over admissible input distributions. GCap evaluates this quantity after replacing the fitted residuals with matched-variance Gaussian noise, whereas Dual-flow approximates the empirical residual distribution through residual resampling and optimizes the input distribution. Their comparison therefore assesses the contribution of distribution-aware capacity estimation while holding the fitted FIR channel fixed.

Like other pairwise measures, channel capacity is conditional on the fitted model and observed variables. For example, if \(X\) drives both \(Y\) and \(Z\), an analysis of \(Y\) and \(Z\) may detect directional dependence induced by the omitted common driver \(X\), despite the absence of a direct \(Y\leftrightarrow Z\) connection. Indirect pathways, feedback, and heterogeneous hemodynamic responses may similarly affect the fitted channel. We evaluate robustness to these factors using simulations with prespecified ground-truth connectivity (Section~\ref{sec:simulation}; Appendix~\ref{app:simulation_details}).

Channel capacity has previously been used to characterize information limits in single neurons and neuronal circuits \citep{mackay1952limiting,ikeda2009capacity,shew2011information,barral2019propagation}. Here, our novelty is to formulate the capacity of a fitted directional interaction between brain regions as a model-based EC measure under general residual distributions. This measure characterizes distribution-dependent communication potential beyond coupling magnitude alone. In this study, the channel is instantiated by a pairwise FIR--residual model within short windows; extension to multivariate autoregressive or other dynamical channel models is a future direction. Because the resulting capacity is conditional on the selected model and observed variables, it should not be interpreted as proof of a direct anatomical connection, established neural causation, or the amount of information actually transmitted in vivo. Before this metric can be translated into a clinically actionable biomarker, rigorous disease-specific validation in independent cohorts will be needed.

\subsection{Baseline EC Methods}
\label{app:baseline}
We selected four representative EC baselines spanning complementary frameworks that remain in current neuroscience and neuroimaging use, supporting their inclusion as contemporary baselines  \citep{zhu2023distinct,mellema2023novel,arab2025whole,murray2024data}. These methods can be applied to the same ROI timeseries and evaluated against the same directed ground truth without auxiliary experimental inputs or anatomical priors. Table~\ref{tab:baseline_comparison} summarizes their main differences.
\begin{table}[h!]
\centering
\caption{Distributional structure and EC estimation principles of the proposed method and evaluated baselines.}
\label{tab:baseline_comparison}
\small
\begin{tabular}{p{0.15\textwidth}p{0.36\textwidth}p{0.39\textwidth}}
\toprule
Method & Distributional structure & EC estimation principle \\
\midrule
Dual-flow
& Residual distribution approximated by empirical samples.
& Incorporates empirical residual shape into channel-capacity estimation while optimizing the input distribution. \\

GCap
& Gaussian residuals; variance estimated from empirical data.
& Computes channel capacity using a matched-variance Gaussian residual model. \\

GC
& Gaussian residuals for conventional parametric inference.
& Compares residual variances of reduced and full autoregressive models to quantify directed predictability. \\

VAR-LiNGAM
& Independent non-Gaussian structural errors; distributional form unspecified.
& Uses non-Gaussianity and independence to identify directed connections. \\

GIMME
& Gaussian signal model under conventional SEM estimation.
& Uses signal covariance structure to select group- and individual-level directed paths. \\
\bottomrule
\end{tabular}
\end{table}

\paragraph{Gaussian Capacity (GCap).}
GCap serves as the within-framework ablation for Dual-flow. Both methods use the same fitted directional FIR channel, input-power constraint, and edge-selection rule. GCap replaces the empirical residual distribution with a matched-variance Gaussian model and computes the corresponding Gaussian-channel capacity, whereas Dual-flow uses empirical residual samples and learns an admissible input distribution. Their comparison therefore tests whether residual distributional shape beyond variance affects EC estimates.

\paragraph{Granger Causality (GC).}
Several GC variants for EC estimation were evaluated by \citet{smith2011network}. Here, we implement the pairwise Granger B$n$ approach, the best-performing GC variant in that benchmark, to provide a direct comparison with our pairwise channel model. This approach uses BIC to select the model order up to a maximum lag of $n$. Throughout this paper, GC refers to this pairwise implementation. For each ordered ROI pair \(X\rightarrow Y\), GC compares a reduced autoregressive model that predicts \(Y\) from its own past with a full model that additionally includes the past of \(X\). The reduction in residual variance quantifies the additional predictive contribution of \(X\) \citep{granger1969investigating,barnett2014mvgc}. The conventional parametric F-test assesses this improvement under Gaussian-error assumptions \citep{seth2015granger}.

\paragraph{VAR-LiNGAM.}
LiNGAM-family methods identify linear causal structure under assumptions of independent non-Gaussian structural errors and acyclicity \citep{shimizu2006linear}. VAR-LiNGAM extends this framework to timeseries data, estimating contemporaneous and lagged directed effects while retaining an acyclic contemporaneous structure \citep{hyvarinen2010estimation}. The specific forms of the structural error distributions are unspecified; their non-Gaussianity and independence enable identification of the directed structure and connection coefficients. We include VAR-LiNGAM as a representative non-Gaussian EC baseline. Our framework uses distributional information differently: it incorporates the empirical residual distribution directly into a channel-capacity EC measure while optimizing the admissible input distribution under a power constraint.

\paragraph{GIMME.}
Group Iterative Multiple Model Estimation (GIMME) is a network-level method based on unified structural equation modeling \citep{gates2012group,sanchezromero2019estimating}. It identifies contemporaneous and lagged connections supported across participants and then adds participant-specific paths. Thus, GIMME estimates group- and individual-level networks, whereas Dual-flow assigns a continuous capacity score to each ordered ROI pair.  Published guidance indicates that GIMME generally works well with 3--20 ROIs, with the required time-series length depending on network size and model complexity \citep{beltz2017network}. The software documentation specifies at least 30 time points per evaluated timeseries, with 60 or more recommended.\footnote{\url{https://github.com/GatesLab/gimme}}

\section{Proofs and Justifications for Theoretical Properties}
\label{app:proofs}

This appendix gives the formal derivations supporting the identities and theoretical properties used in the main text. Throughout, differential entropies are assumed to exist and be finite, the additive noise is independent of admissible inputs, and all distributions are absolutely continuous with respect to Lebesgue measure unless otherwise stated. The fitted residual pool is treated as a finite sample from an unknown continuous residual-noise distribution; empirical resampling is used for Monte Carlo approximation of expectations, not as the definition of the continuous entropy itself.

\subsection{Additive-Channel Capacity Identity}
\label{app:additive_capacity_identity}
Consider the additive FIR vector channel
\begin{align}
    Y=BX+W,
\end{align}
where $B$ is fixed within a temporal window or condition segment and $X$ is independent of $W$. For any admissible input distribution $P_X\in\mathcal P_P$, the mutual information satisfies
\begin{align}
    I(X;Y)
    &= h(Y)-h(Y|X).
\end{align}
Conditioning on $X=x$ gives
\begin{align}
    Y|X=x = Bx+W.
\end{align}
Differential entropy is invariant to deterministic translations, so
\begin{align}
    h(Y|X=x)=h(Bx+W)=h(W).
\end{align}
Averaging over $X$ yields
\begin{align}
    h(Y|X)=\mathbb E_X[h(Y|X=x)]=h(W).
\end{align}
Therefore,
\begin{align}
    I(X;Y)=h(BX+W)-h(W).
\end{align}
Taking the supremum over all admissible input distributions gives Eq.~\ref{eq:additive_capacity_identity}.

\subsection{Proof of Lemma 1: Observer-Flow Entropy Identity}
\label{app:observer_entropy_identity}
Let $q_\phi:\mathbb R^d\rightarrow\mathbb R^d$ be an invertible differentiable flow with Jacobian $J_{q_\phi}(y)$, and let $U\sim\mathcal N(0,I)$ have density $p_U$. The density induced on $Y$ by the observer flow is
\begin{align}
    p_\phi(y)
    =
    p_U(q_\phi(y))\left|\det J_{q_\phi}(y)\right|.
\end{align}
Let $P_Y$ denote the true output distribution with density $p_Y$. The expected observer negative log-likelihood is the cross-entropy from $P_Y$ to $P_\phi$:
\begin{align}
    \mathbb E_{Y\sim P_Y}[-\log p_\phi(Y)]
    &=
    \int p_Y(y)\log\frac{1}{p_\phi(y)}\,dy.
\end{align}
Add and subtract $\log p_Y(y)$:
\begin{align}
    \mathbb E_{P_Y}[-\log p_\phi(Y)]
    &=
    \int p_Y(y)\log\frac{1}{p_Y(y)}\,dy
    +
    \int p_Y(y)\log\frac{p_Y(y)}{p_\phi(y)}\,dy \\
    &=
    h(Y)+\mathrm{KL}(P_Y\|P_\phi).
\end{align}
Thus, if the observer family contains the true density or can approximate it arbitrarily well in KL divergence, the minimized observer loss equals $h(Y)$.

\subsection{Gaussian Capacity as a Special Case}
\label{app:gaussian_recovery}
Assume $W\sim\mathcal N(0,\Sigma_W)$ with $\Sigma_W\succ0$, and let $K_X=\mathrm{Cov}(X)$ for an arbitrary admissible input $X$ satisfying $\mathrm{tr}(K_X)\le dP$. Then
\begin{align}
    \mathrm{Cov}(Y)=BK_XB^\top+\Sigma_W.
\end{align}
Among all distributions with a fixed covariance, the Gaussian distribution maximizes differential entropy. Therefore,
\begin{align}
    h(Y)
    \le
    \frac{1}{2}\log\left((2\pi e)^d\det(BK_XB^\top+\Sigma_W)\right).
\end{align}
Since $W$ is Gaussian,
\begin{align}
    h(W)=\frac{1}{2}\log\left((2\pi e)^d\det\Sigma_W\right).
\end{align}
Using the additive-channel identity,
\begin{align}
    I(X;Y)
    \le
    \frac{1}{2}\log\det\left(I+\Sigma_W^{-1/2}BK_XB^\top\Sigma_W^{-1/2}\right).
\end{align}
The upper bound is achieved by Gaussian $X$ with covariance $\Sigma_X$ solving the constrained optimization problem in Eq.~\ref{eq:gaussian_capacity_matrix}, yielding the classical Gaussian-channel capacity.

\subsection{Entropy-Power Bounds for Matched-Variance Non-Gaussian Residuals}
\label{app:entropy_power}
For the scalar additive channel $Y=X+W$, choose $X\sim\mathcal N(0,P)$ independent of $W$. The entropy power inequality gives
\begin{align}
    N(X+W)\ge N(X)+N(W)=P+N(W).
\end{align}
Therefore,
\begin{align}
    I(X;Y)
    &=h(X+W)-h(W)\\
    &=\frac{1}{2}\log\frac{N(X+W)}{N(W)}\\
    &\ge \frac{1}{2}\log\left(1+\frac{P}{N(W)}\right).
\end{align}
Because capacity is the supremum over admissible input distributions, the same
quantity provides an achievable-rate lower bound on $C(W)$. Moreover, Gaussian
distributions maximize entropy at fixed variance, so
$N(W)\le\sigma_W^2$, with equality only when $W$ is Gaussian. This yields
Eq.~\ref{eq:nongaussian_capacity_gap} and shows that matched residual variance
does not, in general, imply matched achievable information rate.

\paragraph{Extension to the FIR channel.}
To obtain the bounds used in the controlled matched-variance experiment, we consider the finite-dimensional additive FIR channel in its vector form
\begin{align}
    Y = BX + W,
\end{align}
where \(B\in\mathbb R^{d\times d}\) is the finite-dimensional FIR convolution operator induced by the fixed filter \(b\), the admissible input satisfies
\begin{align}
    \mathbb E\|X\|_2^2 \le dP,
\end{align}
and the residual noise \(W\) has matched per-sample variance \(\sigma_W^2\). In the Gaussian-surrogate model, the residual covariance is taken to be \(\Sigma_W=\sigma_W^2 I_d\).

\paragraph{Gaussian surrogate capacity.}
By the Gaussian special case in Appendix~\ref{app:gaussian_recovery}, replacing the true residual law by a Gaussian distribution with the same variance gives
\begin{align}
    C_{\mathrm{G}}^{\mathrm{FIR}}
    &=
    \frac{1}{2d}
    \max_{\Sigma_X\succeq 0,\;\mathrm{tr}(\Sigma_X)\le dP}
    \log\det\!\left(
        I_d+\frac{1}{\sigma_W^2}B\Sigma_X B^\top
    \right).
    \label{eq:fir_gaussian_surrogate_capacity}
\end{align}
Equivalently, if \(\lambda_i\) are the squared singular values of \(B\) and \(p_i^\star\) are the corresponding water-filling allocations under the power constraint \(\sum_i p_i^\star \le dP\), then
\begin{align}
    C_{\mathrm{G}}^{\mathrm{FIR}}
    =
    \frac{1}{2d}
    \sum_i
    \log\!\left(
        1+\frac{p_i^\star \lambda_i}{\sigma_W^2}
    \right).
    \label{eq:fir_gaussian_surrogate_capacity_svd}
\end{align}
Because \(B\), \(P\), and \(\sigma_W^2\) are fixed across matched-variance residual conditions, \(C_{\mathrm{G}}^{\mathrm{FIR}}\) is fixed across residual distributions.

\paragraph{Lower bound.}
Let \(\Sigma_X^\star\) denote the Gaussian water-filling input covariance attaining Eq.~\ref{eq:fir_gaussian_surrogate_capacity}. A lower bound for the same FIR model follows from the vector entropy power inequality applied to the admissible Gaussian input \(X_G\sim\mathcal N(0,\Sigma_X^\star)\):
\begin{align}
    C^{\mathrm{FIR}}(W)
    &\ge
    \frac{1}{2}
    \log\!\left(
        1+\frac{\det(B\Sigma_X^\star B^\top)^{1/d}}{N(W)}
    \right),
    \label{eq:fir_lower_bound_fullrank}
\end{align}
where the vector entropy power of \(W\) is
\begin{align}
    N(W)
    =
    \frac{1}{2\pi e}
    \exp\!\left(\frac{2}{d}h(W)\right).
    \label{eq:vector_entropy_power}
\end{align}
In practice, \(B\Sigma_X^\star B^\top\) may be rank-deficient because only a subset of singular modes receive nonzero water-filling power. Let
\begin{align}
    \mathcal I_\star = \{i:\lambda_i>0,\;p_i^\star>0\},
    \qquad
    r = |\mathcal I_\star|.
\end{align}
Restricting the entropy-power argument to the active output subspace yields the implementable lower bound used in the simulation:
\begin{align}
    C_{\mathrm{LB}}^{\mathrm{FIR}}(W)
    &=
    \frac{r}{2d}
    \log\!\left(
        1+\frac{
            \left(
                \prod_{i\in\mathcal I_\star}\lambda_i p_i^\star
            \right)^{1/r}
        }{N(W)}
    \right).
    \label{eq:fir_lower_bound_active}
\end{align}
For i.i.d.\ residuals with per-sample differential entropy \(h(w)\), Eq.~\ref{eq:vector_entropy_power} reduces to
\begin{align}
    N(W)=\frac{1}{2\pi e}\exp(2h(w)).
\end{align}

\paragraph{Upper bound.}
An upper bound for the same FIR model follows from the fact that, for fixed covariance, the Gaussian distribution maximizes differential entropy. Let \(h_G\) denote the per-sample entropy of a matched Gaussian residual with variance \(\sigma_W^2\):
\begin{align}
    h_G
    =
    \frac{1}{2}\log(2\pi e\,\sigma_W^2).
\end{align}
Then
\begin{align}
    C^{\mathrm{FIR}}(W)
    &\le
    C_{\mathrm{G}}^{\mathrm{FIR}} + h_G - h(w),
    \label{eq:fir_upper_bound}
\end{align}
where \(h(w)\) is the per-sample differential entropy of the true residual law. Equivalently, in vector form,
\begin{align}
    C^{\mathrm{FIR}}(W)
    &\le
    C_{\mathrm{G}}^{\mathrm{FIR}}
    +
    \frac{1}{d}
    \left[
        \frac{1}{2}\log\!\bigl((2\pi e)^d\det(\sigma_W^2 I_d)\bigr)
        - h(W)
    \right].
\end{align}

\paragraph{Implications for matched-variance residuals.}
For Gaussian residuals, \(N(W)=\sigma_W^2\) and \(h(w)=h_G\), so the upper bound in Eq.~\ref{eq:fir_upper_bound} collapses exactly to the Gaussian-surrogate capacity:
\begin{align}
    C_{\mathrm{UB}}^{\mathrm{FIR}}(W_{\mathrm{Gaussian}})
    =
    C_{\mathrm{G}}^{\mathrm{FIR}}.
\end{align}
By contrast, the lower bound in Eq.~\ref{eq:fir_lower_bound_active} is generally not tight even in the Gaussian case, so typically
\begin{align}
    C_{\mathrm{LB}}^{\mathrm{FIR}}(W_{\mathrm{Gaussian}})
    \le
    C_{\mathrm{G}}^{\mathrm{FIR}}
    =
    C_{\mathrm{UB}}^{\mathrm{FIR}}(W_{\mathrm{Gaussian}}).
\end{align}
For matched-variance non-Gaussian residuals, $C_{\mathrm{G}}^{\mathrm{FIR}}$ remains fixed, whereas the lower and upper bounds depend on residual entropy through $N(W)$ and $h(w)$. Thus residual distributional structure can alter the range of achievable information rates even when the FIR channel, input-power constraint, and residual variance are
held fixed.

\subsection{Game-Gap Diagnostic}
\label{app:game_gap}

Although the main text does not rely on a theory-heavy saddle-point claim, the dual-flow objective can be monitored with an empirical game-gap diagnostic.

\begin{proposition}[Game gap for the dual-flow objective]
\label{prop:game_gap}
Let
\begin{align}
    V_{\Theta,\Phi}
    =
    \sup_{\theta\in\Theta}\inf_{\phi_Y\in\Phi}\mathcal L_Y(\theta,\phi_Y)
\end{align}
denote the signal-plus-noise value of the flow-restricted game. For a trained pair \((\widehat\theta,\widehat\phi_Y)\), define the generator and observer regrets
\begin{align}
    r_{\mathrm{gen}}
    &=
    \sup_{\theta\in\Theta}\mathcal L_Y(\theta,\widehat\phi_Y)
    -
    \mathcal L_Y(\widehat\theta,\widehat\phi_Y), \\
    r_{\mathrm{obs}}
    &=
    \mathcal L_Y(\widehat\theta,\widehat\phi_Y)
    -
    \inf_{\phi_Y\in\Phi}\mathcal L_Y(\widehat\theta,\phi_Y).
\end{align}
Define the game gap as
\begin{align}
    \mathrm{Gap}_Y = r_{\mathrm{gen}} + r_{\mathrm{obs}}.
\end{align}
Then
\begin{align}
    \left|
    \mathcal L_Y(\widehat\theta,\widehat\phi_Y)
    -V_{\Theta,\Phi}
    \right|
    \le
    \max\{r_{\mathrm{gen}},r_{\mathrm{obs}}\}
    \le
    \mathrm{Gap}_Y.
    \label{eq:game_gap_bound}
\end{align}
Thus a small game gap indicates that neither player can substantially improve the objective by unilateral re-optimization, and bounds the remaining optimization error relative to the flow-restricted saddle-point value.
\end{proposition}

\paragraph{Proof.}
Let
\begin{align}
    V_{\Theta,\Phi}
    =
    \sup_{\theta\in\Theta}\inf_{\phi\in\Phi}\mathcal L_Y(\theta,\phi)
\end{align}
be the max--min value of the signal-plus-noise flow game. For any fixed observer \(\widehat\phi_Y\), weak duality gives
\begin{align}
    V_{\Theta,\Phi}
    \le
    \sup_{\theta\in\Theta}\mathcal L_Y(\theta,\widehat\phi_Y).
\end{align}
By the definition of the generator regret,
\begin{align}
    \sup_{\theta\in\Theta}\mathcal L_Y(\theta,\widehat\phi_Y)
    =
    \mathcal L_Y(\widehat\theta,\widehat\phi_Y)+r_{\mathrm{gen}}.
\end{align}
Therefore
\begin{align}
    V_{\Theta,\Phi}
    \le
    \mathcal L_Y(\widehat\theta,\widehat\phi_Y)+r_{\mathrm{gen}}.
    \label{eq:app_game_upper}
\end{align}
Similarly, for any fixed generator \(\widehat\theta\),
\begin{align}
    \inf_{\phi\in\Phi}\mathcal L_Y(\widehat\theta,\phi)
    \le
    V_{\Theta,\Phi}.
\end{align}
By the definition of the observer regret,
\begin{align}
    \inf_{\phi\in\Phi}\mathcal L_Y(\widehat\theta,\phi)
    =
    \mathcal L_Y(\widehat\theta,\widehat\phi_Y)-r_{\mathrm{obs}}.
\end{align}
Thus
\begin{align}
    \mathcal L_Y(\widehat\theta,\widehat\phi_Y)-r_{\mathrm{obs}}
    \le
    V_{\Theta,\Phi}.
    \label{eq:app_game_lower}
\end{align}
Combining Eqs.~\ref{eq:app_game_upper} and \ref{eq:app_game_lower} gives
\begin{align}
    \mathcal L_Y(\widehat\theta,\widehat\phi_Y)-r_{\mathrm{obs}}
    \le
    V_{\Theta,\Phi}
    \le
    \mathcal L_Y(\widehat\theta,\widehat\phi_Y)+r_{\mathrm{gen}}.
\end{align}
Therefore
\begin{align}
    \left|
    \mathcal L_Y(\widehat\theta,\widehat\phi_Y)-V_{\Theta,\Phi}
    \right|
    \le
    \max\{r_{\mathrm{gen}},r_{\mathrm{obs}}\}
    \le
    r_{\mathrm{gen}}+r_{\mathrm{obs}}.
\end{align}
This proves Eq.~\ref{eq:game_gap_bound}. The same reasoning applies to the noise-only observer objective after removing the generator player, where the relevant optimization error is the suboptimality gap
\begin{align}
    \mathcal L_W(\widehat\phi_W)-\inf_{\phi_W\in\Phi}\mathcal L_W(\phi_W).
\end{align}

In practice, this diagnostic can be estimated by restarting or continuing optimization of one player while holding the other fixed.

\subsection{Detailed Error Decomposition}
\label{app:error_decomposition}
The error decomposition in Eq.~\ref{eq:error_decomposition} separates channel and residual estimation, generator restriction, observer approximation, optimization, and Monte Carlo error. It is an accounting of approximation sources rather than a finite-sample guarantee. Throughout this section, differential entropy is defined for the underlying continuous distributions. Empirical residual resampling is used to approximate expectations and is not assigned a differential entropy.

Let \(C\) denote the population capacity of the additive FIR channel with operator \(B\) and continuous residual-noise law \(P_W\). For \(X_\theta=\mathrm{Norm}_P(g_\theta(Z_1))\), define the generator-restricted capacity
\begin{align}
    C_\Theta
    =
    \sup_{\theta\in\Theta}
    \left\{h(BX_\theta+W)-h(W)\right\},
    \qquad W\sim P_W.
\end{align}
The generator-restriction error is
\begin{align}
    \varepsilon_{\mathrm{gen}}=C-C_\Theta.
\end{align}
This term measures the loss from restricting the admissible input distributions to the generator family.

Using the population losses \(\mathcal L_Y\) and \(\mathcal L_W\), define the flow-restricted entropy-difference objective
\begin{align}
    V_{\Theta,\Phi}
    =
    \sup_{\theta\in\Theta}\inf_{\phi_Y\in\Phi}
    \mathcal L_Y(\theta,\phi_Y)
    -
    \inf_{\phi_W\in\Phi}\mathcal L_W(\phi_W).
\end{align}
For \(Y_\theta=BX_\theta+W\), Lemma~1 gives
\begin{align}
    \inf_{\phi_Y\in\Phi}\mathcal L_Y(\theta,\phi_Y)
    &=h(Y_\theta)+a_\Phi(\theta),\\
    a_\Phi(\theta)
    &=\inf_{\phi_Y\in\Phi}
    \mathrm{KL}(P_{Y_\theta}\|P_{\phi_Y}),\\
    \inf_{\phi_W\in\Phi}\mathcal L_W(\phi_W)
    &=h(W)+b_\Phi,\\
    b_\Phi
    &=\inf_{\phi_W\in\Phi}
    \mathrm{KL}(P_W\|P_{\phi_W}).
\end{align}
Consequently, the observer-approximation error satisfies
\begin{align}
    \varepsilon_{\mathrm{obs}}
    :=
    |V_{\Theta,\Phi}-C_\Theta|
    \le
    \sup_{\theta\in\Theta}a_\Phi(\theta)+b_\Phi.
\end{align}

Let \(\widehat B\) be the fitted FIR operator and let \(\widehat P_W\) denote the distribution induced by resampling the fitted residual pool. Conditional on the observed data, define the empirical-resampling losses
\begin{align}
    \widetilde{\mathcal L}_Y(\theta,\phi_Y)
    &=
    \mathbb E_{Z_1,\widehat W}
    \left[-\log p_{\phi_Y}
    \left(\widehat B X_\theta+\widehat W\right)\right],\\
    \widetilde{\mathcal L}_W(\phi_W)
    &=
    \mathbb E_{\widehat W}
    \left[-\log p_{\phi_W}(\widehat W)\right],
    \qquad \widehat W\sim\widehat P_W.
\end{align}
These are expected log-density losses evaluated on resampled data, not differential entropies of the empirical distribution. Define their optimized difference as
\begin{align}
    \widetilde V_{\Theta,\Phi}
    =
    \sup_{\theta\in\Theta}\inf_{\phi_Y\in\Phi}
    \widetilde{\mathcal L}_Y(\theta,\phi_Y)
    -
    \inf_{\phi_W\in\Phi}
    \widetilde{\mathcal L}_W(\phi_W).
\end{align}
The channel and residual approximation term is
\begin{align}
    \varepsilon_{\mathrm{resid}}
    =
    |\widetilde V_{\Theta,\Phi}-V_{\Theta,\Phi}|.
\end{align}
This term captures the effect of replacing the population channel and residual expectations with the fitted FIR operator and finite residual pool. Increasing the number of resampled training points does not, by itself, eliminate this discrepancy.

For trained parameters \((\widehat\theta,\widehat\phi_Y,\widehat\phi_W)\), let
\begin{align}
    \widetilde C_{\mathrm{train}}
    =
    \widetilde{\mathcal L}_Y(\widehat\theta,\widehat\phi_Y)
    -
    \widetilde{\mathcal L}_W(\widehat\phi_W).
\end{align}
Applying the argument of Proposition~\ref{prop:game_gap} to the empirical-resampling losses gives
\begin{align}
    |\widetilde C_{\mathrm{train}}-\widetilde V_{\Theta,\Phi}|
    \le
    \mathrm{Gap}_{\mathrm{game}},
\end{align}
where \(\mathrm{Gap}_{\mathrm{game}}\) is the sum of the generator regret, signal-plus-noise observer regret, and noise-only observer suboptimality for these losses.

Finally, define
\begin{align}
    \varepsilon_{\mathrm{MC}}
    =
    |\widehat C_{\Theta,\Phi}-\widetilde C_{\mathrm{train}}|
\end{align}
as the error from evaluating the trained empirical-resampling objectives using finite Monte Carlo samples. Adding and subtracting the intermediate quantities yields
\begin{align}
    |\widehat C_{\Theta,\Phi}-C|
    \le
    \varepsilon_{\mathrm{resid}}
    +
    \varepsilon_{\mathrm{gen}}
    +
    \varepsilon_{\mathrm{obs}}
    +
    \mathrm{Gap}_{\mathrm{game}}
    +
    \varepsilon_{\mathrm{MC}}.
\end{align}
This decomposition assumes that the displayed quantities are finite. It does not establish that the terms are small or provide convergence rates; it identifies the distinct sources of discrepancy between the finite-sample estimate and population capacity.

\section{Algorithm, Training Sensitivity, and Controlled Validation of the Dual-Flow Estimator}
\subsection{Algorithm and Implementation Details}
\label{app:algorithm}
\begin{algorithm}[H]
\caption{Dual-flow empirical channel-capacity estimation}
\label{alg:dual_flow_capacity}
\begin{algorithmic}[1]
\Require Source signal $x_{1:T}$, target signal $y_{1:T}$, FIR order $k$, power constraint $P$, training-sequence length $L$, generator $g_\theta$, observer flows $q_{\phi_Y}$ and $q_{\phi_W}$
\State Fit FIR coefficients $\widehat b$ from $x$ to $y$ by least squares.
\State Compute residuals $\widehat w_t = y_t-\sum_{\ell=0}^{k-1}\widehat b_\ell x_{t-\ell}$ for $t=k,\ldots,T$.
\Repeat
    \State Sample a latent sequence $Z_1\sim\mathcal N(0,I_L)$ of length $L$.
    \State Sample a residual-noise sequence $W$ of length $L$ by uniformly resampling with replacement from $\{\widehat w_t\}_{t=k}^{T}$.
    \State Generate admissible inputs $X_\theta=\mathrm{Norm}_P(g_\theta(Z_1))$.
    \State Compute signal-plus-noise outputs $Y_\theta=BX_\theta+W$.
    \State Update observer $q_{\phi_Y}$ to minimize $-\log p_{\phi_Y}(Y_\theta)$.
    \State Update noise observer $q_{\phi_W}$ to minimize $-\log p_{\phi_W}(W)$.
    \State Update generator $g_\theta$ to maximize the signal-plus-noise objective $\mathcal L_Y(\theta,\phi_Y)$.
    \State Estimate capacity $\widehat C=\widehat h(Y_\theta)-\widehat h(W)$.
\Until{the exponential moving average of $\widehat C$ reaches stationarity}
\State \Return $\widehat C$
\end{algorithmic}
\end{algorithm}

\paragraph{Flow architecture.}
Both the generator and observer are implemented using three pairs of 1D Real NVP coupling layers (Fig.~\ref{fig:fig1}b). Each coupling layer partitions the input into frozen and transformed components using a binary mask, and paired layers use complementary masks so that all dimensions are transformed. The transformed component is updated by scale and translation networks, each implemented with two one-dimensional convolutional layers and an intermediate ReLU nonlinearity. In each observer coupling layer, the sum of the scale-network outputs gives the log-determinant of the layer Jacobian; summing these terms across all observer layers yields $\log|\det J_{q_\phi}(y)|$ in Eq.~\ref{eq:observer_loss}. To stabilize training, scale outputs are softly clipped as $s\mapsto 2\tanh(s/2)$.

\paragraph{Training and optimization.}
At each training iteration, a latent sequence \(Z_1\sim\mathcal N(0,I_L)\) of length \(L\) and \(L\) residual-noise samples drawn by uniform resampling from the fitted residual pool are propagated through the generator, FIR channel, and observer. We alternate observer updates that minimize negative log-likelihood with generator updates that maximize the signal-plus-noise entropy estimate. Gradients are computed by automatic differentiation and parameters are optimized with {\it Adam}. Training stops when the exponential moving average of the paired entropy-difference estimate reaches stationarity. Sensitivity to the generated training-sequence length is reported in Appendix~\ref{app:L_sensitivity}.

\paragraph{Residual-resampling approximation.} In empirical data, the true residual law $P_W$ is unknown, so we approximate expectations under this law by uniformly resampling with replacement from the fitted residual pool. The quality of this approximation depends on the size and representativeness of the pool and the adequacy of the fitted FIR model. Under our indexing, an observed segment of length \(T\) contributes \(T-k+1\) fitted residuals, from which \(L\) residual samples are drawn with replacement at each training iteration. Thus, \(T\) determines the amount of empirical residual information available, whereas \(L\) determines the generated training-sequence length used for Monte Carlo estimation; increasing \(L\) does not add information beyond the observed residual pool. Independent resampling captures the empirical marginal residual distribution but does not preserve temporal dependence, and therefore relies on treating residuals as approximately independent draws from a stable within-segment distribution. Strong temporal dependence or within-segment changes in the residual distribution can compromise this approximation.

\subsection{{Sensitivity to Training Sequence Length}}
\label{app:L_sensitivity}
We evaluated sensitivity to the generated training-sequence length \(L\), using \(L=1024\) as the default. We tested \(L\in\{256,512,1024,2048,4096\}\) using 50 random initializations for each of four residual distributions (Gaussian, Uniform, Student's \(t\), and Exponential), yielding 1{,}000 runs in total. Across all conditions, mean capacity estimates differed from their corresponding \(L=1024\) values by at most 4.30\%, and all mean shifts were no larger than the corresponding across-initialization SD at \(L=1024\). All 1{,}000 runs completed without numerical failure and produced finite estimates, indicating limited sensitivity to \(L\) within the tested range (Table~\ref{tab:T_sensitivity}). The Real-NVP architecture with three coupling-layer pairs and the EMA-based convergence criterion were fixed across all experiments rather than tuned to individual datasets. A broader architecture-depth ablation was not performed and is acknowledged as a limitation.

\begin{table}[h!]
\centering
\caption{Capacity estimates across training sequence lengths. Values are mean \(\pm\) SD across 50 random initializations. Max. \(\Delta\) is the largest relative deviation from the \(L=1024\) value.}
\label{tab:T_sensitivity}

\begin{tabular}{lccccc c}
\toprule
Distribution & \(L{=}256\) & \(L{=}512\) & \(L{=}1024\) & \(L{=}2048\) & \(L{=}4096\) & Max. \(\Delta\) \\
\midrule
Gaussian      & .128$\pm$.006 & .129$\pm$.008 & .131$\pm$.007 & .132$\pm$.011 & .134$\pm$.011 & 2.25\% \\
Uniform       & .236$\pm$.011 & .238$\pm$.010 & .237$\pm$.012 & .239$\pm$.012 & .238$\pm$.011 & 1.11\% \\
Student's \(t\) & .184$\pm$.011 & .181$\pm$.010 & .184$\pm$.006 & .185$\pm$.013 & .190$\pm$.017 & 3.23\% \\
Exponential   & .337$\pm$.030 & .346$\pm$.034 & .332$\pm$.038 & .344$\pm$.029 & .341$\pm$.037 & 4.30\% \\
\bottomrule
\end{tabular}
\end{table}
\subsection{Controlled Validation of Estimator Properties}
\label{app:controlled_validation}

\begin{figure}[h!]
    \centering
    \includegraphics[width=0.98\linewidth]{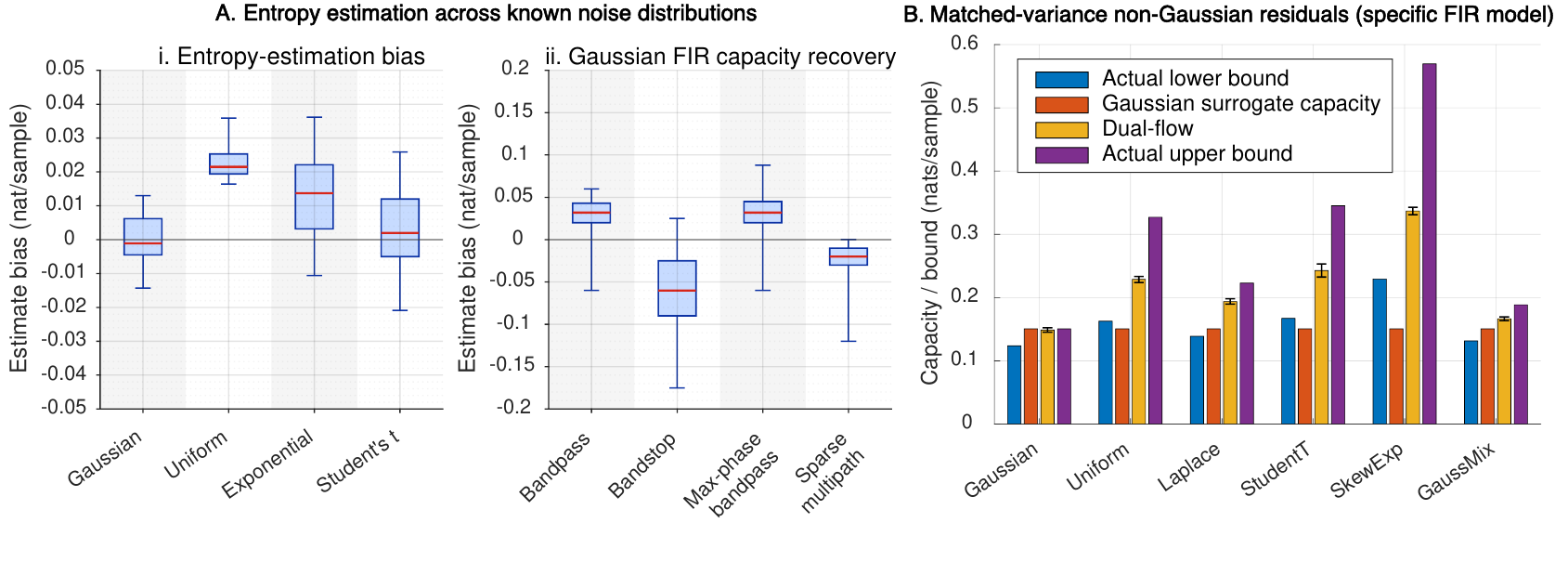}
    \caption{\textbf{Controlled validation of Dual-flow.} \textbf{A}, Validation in analytically tractable settings. \textbf{(i)} Entropy-estimation bias relative to ground truth across known noise distributions over 60 runs. \textbf{(ii)} Capacity-estimation bias relative to analytic Gaussian FIR capacity across representative channels over 60 runs. \textbf{B}, Matched-variance non-Gaussian residuals in a fixed FIR channel. Bars show the actual lower bound, Gaussian surrogate capacity, Dual-flow estimate, and actual upper bound for residual distributions with matched variance. Error bars denote mean \(\pm\) s.e.m. where applicable.
} \label{fig:fig_sim}
\end{figure}

\paragraph{Entropy estimation and Gaussian-capacity recovery.}
We first evaluated the estimator in analytically tractable settings where the target entropy or channel capacity is known. For entropy validation, we set the FIR coefficients to zero so that the channel output consisted only of sampled noise. Noise samples were drawn from Gaussian, Uniform, Exponential, and Student's \(t\) distributions, with \(\nu=3\) for Student's \(t\). For each distribution, we performed \(N=60\) independent training runs and compared the observer-flow entropy estimate with the ground-truth differential entropy. Across all four distributions, the observer accurately recovered the known entropy, with small estimation bias across independent runs (Fig.~\ref{fig:fig_sim}A-i). For Gaussian-capacity validation, we then evaluated the full generator--observer game in FIR channels with known analytic capacity. We tested four representative channels: a bandpass filter, a bandstop filter, a maximum-phase version of the bandpass filter obtained by time reversal, and a maximum-phase sparse multipath filter with coefficients \(b=(0,0,0,0,0,0,0.1,0,0.3,0,0,0,0,0,0,0.4)\). The bandpass and bandstop filters were constructed using second-order Butterworth filters with passband or stopband between 0.3 and 0.8 times the Nyquist frequency and truncated to the first 16 impulse-response coefficients. Across these representative channel structures, the full Dual-flow estimator recovered the corresponding analytic Gaussian FIR capacities with small estimation bias over 60 independent runs (Fig.~\ref{fig:fig_sim}A-ii). Together, Fig.~\ref{fig:fig_sim}A-i--ii validates observer-based entropy estimation and generator-based capacity maximization against known ground truth before application to empirical non-Gaussian residuals.

\paragraph{Matched-variance FIR validation.}
We next evaluated the theoretical prediction that residual distributions with identical variance can nevertheless support different achievable information rates because of distributional structure beyond second-order statistics. We considered additive FIR channels with fixed filter \(B\), fixed input-power constraint \(P\), and matched residual variance \(\sigma_W^2\), while varying only the residual distribution; the corresponding Gaussian-surrogate capacity and entropy-dependent theoretical bounds are derived in Appendix~\ref{app:entropy_power}. Because \(B\), \(P\), and \(\sigma_W^2\) are fixed, the Gaussian-surrogate capacity is identical across residual laws, whereas the entropy-power lower bound and entropy-based upper bound vary with residual entropy, as shown in Fig.~\ref{fig:fig_sim}B. Dual-flow estimates likewise varied across residual distributions and tracked these entropy-dependent differences rather than remaining fixed at the Gaussian-surrogate value (Fig.~\ref{fig:fig_sim}B). In particular, for several non-Gaussian residual laws, including Uniform, Student's \(t\), and skewed-Exponential residuals, the Gaussian surrogate fell below the theoretical lower bound, demonstrating that a matched-variance Gaussian approximation can underestimate achievable information rates (Fig.~\ref{fig:fig_sim}B). In contrast, Dual-flow remained within the theoretical lower and upper bounds for all non-Gaussian residual cases and agreed with the Gaussian-surrogate capacity in the Gaussian case (Fig.~\ref{fig:fig_sim}B). These controlled matched-variance simulations therefore show that residual distribution shape alone can alter achievable information rates even when the channel filter, input power, and residual variance are fixed, and that Dual-flow captures these entropy-dependent differences beyond variance-based Gaussian capacity.


\section{Implementation Details/Empirical Data Analysis}
\label{app:implementation_details}

\subsection{Brain Datasets and Gaussianity Assessment}
\label{app:gaussianity_analysis}
We evaluated Gaussianity in three datasets spanning human task-fMRI, human simultaneous EEG--fMRI, and rodent LFP--fMRI, chosen to sample distinct recording modalities, temporal scales, and experimental contexts. For the human task-fMRI and EEG--fMRI datasets, the analysis focused on observed signal distributions. For the rat LFP--fMRI dataset, we additionally examined fitted local channel residuals to test whether residual distributions can deviate from Gaussianity even when the corresponding BOLD signals appear approximately Gaussian at the whole-scan level. Across datasets, departures from Gaussianity were summarized primarily using Anderson--Darling (A--D) statistics relative to matched Gaussian references, with larger values indicating stronger deviations from Gaussianity.

\paragraph{Human HCP motor-task fMRI.}
To evaluate Gaussianity in task-evoked human fMRI, we used the motor-task dataset from the Human Connectome Project test--retest cohort \citep{barch2013function}. The minimally preprocessed 3T fMRI data (\(TR=0.72\) s) included 45 participants, each completing two motor-task runs. The task paradigm comprised visually cued movements of the left foot, right foot, left hand, right hand, and tongue, together with interleaved resting blocks \citep{barch2013function}. Task-specific activation maps were estimated using FSL FEAT following the HCP motor-task design and standard preprocessing choices commonly used for this dataset \citep{barch2013function}. ROI time series were extracted from validated task-activated regions in motor, somatosensory, and cerebellar areas, whose spatial patterns were consistent with canonical motor-task activation reported previously \citep{barch2013function,rizzolatti2001cortical,manto2012consensus}. For the Gaussianity analysis, we evaluated ROI-level fMRI signals from the task conditions and matched rest blocks.

\paragraph{Human simultaneous EEG--fMRI.}
We further analyzed {the eyes-open resting-state runs of} a simultaneous EEG--fMRI resting-state dataset 
\citep{ds006040}. The dataset originally contained 28 participants and included concurrent EEG, fMRI, and diffusion-weighted imaging; after excluding subjects with interrupted EEG acquisition or substantial motion artifacts, the final sample consisted of 22 subjects \citep{ds006040}. fMRI data were acquired on a 3T Siemens Prisma scanner with \(TR=2.0\) s, and EEG was recorded concurrently using a 64-channel MR-compatible Brain Products system at 5000 Hz \citep{ds006040}. For EEG processing, we used the derivative data with MR-related cardiac artifacts already removed by the dataset authors, followed by standard preprocessing including rereferencing, ICA-based artifact removal, detrending, and source reconstruction with eLORETA on the fsaverage template \citep{ds006040,pascualmarqui2007eloreta,gramfort2013mne}. Source-space signals were extracted from Schaefer-atlas 100 ROIs \citep{schaefer2018local}, and subsequent analysis focused on default mode network regions to match the resting-state fMRI analysis. In contrast to frequency-specific analyses, here we used broadband source-space EEG signals for the Gaussianity diagnosis. We then evaluated Gaussianity for both the fMRI ROI signals and the broadband EEG source signals.

\paragraph{Rat concurrent LFP--fMRI.}
We used a rat dataset with concurrent 9.4T fMRI and LFP recordings from the left and right primary somatosensory cortex (S1), which served as the two ROIs, to assess Gaussianity in both hemodynamic and electrophysiological signals. The data were acquired under dexmedetomidine (DMED) anesthesia and consisted of repeated resting-state scans with fMRI repetition time \(TR=0.5\) s. fMRI preprocessing followed standard procedures in SPM12, including motion correction, spatial smoothing, global signal regression, and temporal band-pass filtering (0.01--0.1 Hz). For the electrophysiology data, gradient-related artifacts were removed, recordings were aligned to the fMRI frame structure, and broadband LFP power was computed after denoising and low-pass filtering. We also computed band-limited power (BLP) in six frequency bands: delta (1--4 Hz), theta (4--8 Hz), alpha (8--14 Hz), low beta (14--25 Hz), high beta (25--40 Hz), and gamma (40--100 Hz) \citep{pan2011broadband,thompson2013neural}. Quality control and ROI definition for the bilateral S1 fMRI signals followed prior work on this dataset \citep{zhang2020relationship,pan2011broadband,thompson2013neural}. For the Gaussianity analysis, we examined the resting BOLD signals and corresponding LFP broadband power and BLP timeseries.

To further assess whether approximate scan-level Gaussianity can mask local distributional deviations relevant to directed-connectivity estimation, we also fit local FIR channels to the rat resting-state BOLD signals in sliding windows and computed residual samples as Eq.~\ref{eq:channel_residual}. We then quantified Gaussianity of these window-specific residuals. This residual analysis was performed only for the rat BOLD data and was intended to test whether local channel residuals can exhibit stronger non-Gaussianity than the corresponding whole-scan BOLD signals.

\begin{figure}[h!]
    \begin{subfigure}{0.48\linewidth}
        \centering
        \includegraphics[width=\linewidth]{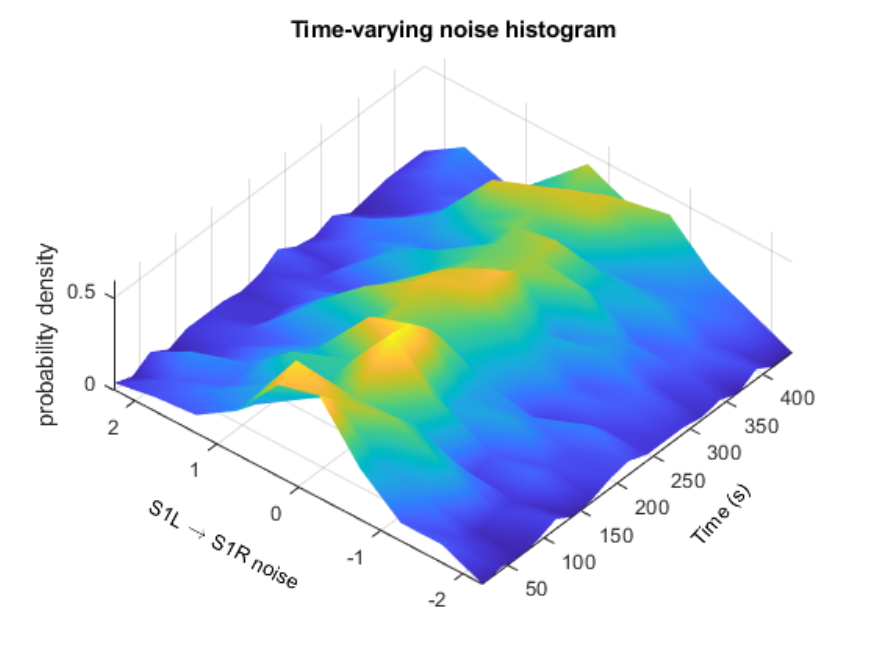}
        \caption{}
        \label{fig:sw_dist}
    \end{subfigure}
    \begin{subfigure}{0.48\linewidth}
        \centering
        \includegraphics[width=\linewidth]{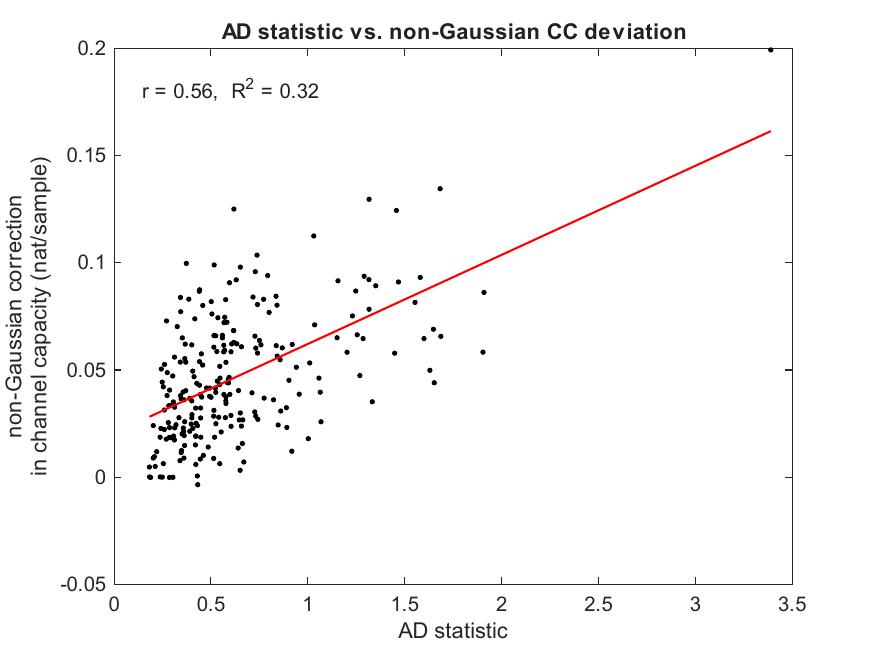}
        \caption{}
        \label{fig:sw_cc}
    \end{subfigure}
    \caption{\textbf{Sliding-window analysis of the LFP--BOLD dataset.} (a) Time-resolved noise histograms in a representative BOLD recording reveal time-variability of the distribution and the deviations from Gaussianity. (b) The difference between GCap and Dual-flow estimates correlates with the Anderson–Darling (AD) statistic of the windowed time series ({Pearson $r = 0.56$}). Each dot represents the results computed for a window from a BOLD or an LFP band-limited power timeseries.}
    \label{fig:lfp_sw}
\end{figure}

Fig.~\ref{fig:sw_dist} illustrates temporal variation in the fitted residual distribution of a representative BOLD recording. Separately, across windows from BOLD and LFP band-limited power time series, larger signal-level A--D statistics were associated with larger differences between GCap and Dual-flow capacity estimates ({Pearson $r = 0.56$}; Fig.~\ref{fig:sw_cc}). This association links stronger signal-level departures from Gaussianity to greater disagreement between Gaussian and distribution-aware EC estimates.

\subsection{Task-Versus-Rest Sensitivity Analysis in Tongue-Motion fMRI}
\label{app:empirical_application}
We analyzed the first tongue-motion trial and matched resting-fixation period from acquisition run~1 for each of the 44 participants with usable data from the 45-participant HCP test--retest cohort. Each segment contained 16 fMRI time points (approximately 12~s) and was analyzed separately without additional sliding windows. GIMME was not evaluated because these segments were shorter than its documented minimum time-series length (Appendix~\ref{app:baseline}). The optimal FIR model order was estimated independently for each directed ROI pair and trial using the corrected Akaike information criterion (AICc) \citep{hurvich1989regression}.

For each directed ROI pair, we fit the local FIR model (Eq.~\ref{eq:channel}) within each task or rest trial, and estimated four directed-connectivity measures from the same ROI time series: (i) Dual-flow, which treats the fitted residuals as empirical samples from an unknown continuous residual-noise distribution; (ii) Gaussian capacity (GCap), which uses the same fitted FIR filter but replaces residual uncertainty by a matched Gaussian model based on residual second-order statistics; (iii) pairwise Granger causality as a directed-prediction baseline; and (iv) VAR-LiNGAM, which uses non-Gaussian innovations for structural identification in a vector-autoregressive model. This parallel construction allowed us to compare whether explicitly modeling residual distribution shape improves sensitivity to task-evoked directed interactions beyond Gaussian capacity and Granger-causal baselines.

\begin{figure}[h!]
    \centering
    \includegraphics[width=380pt]{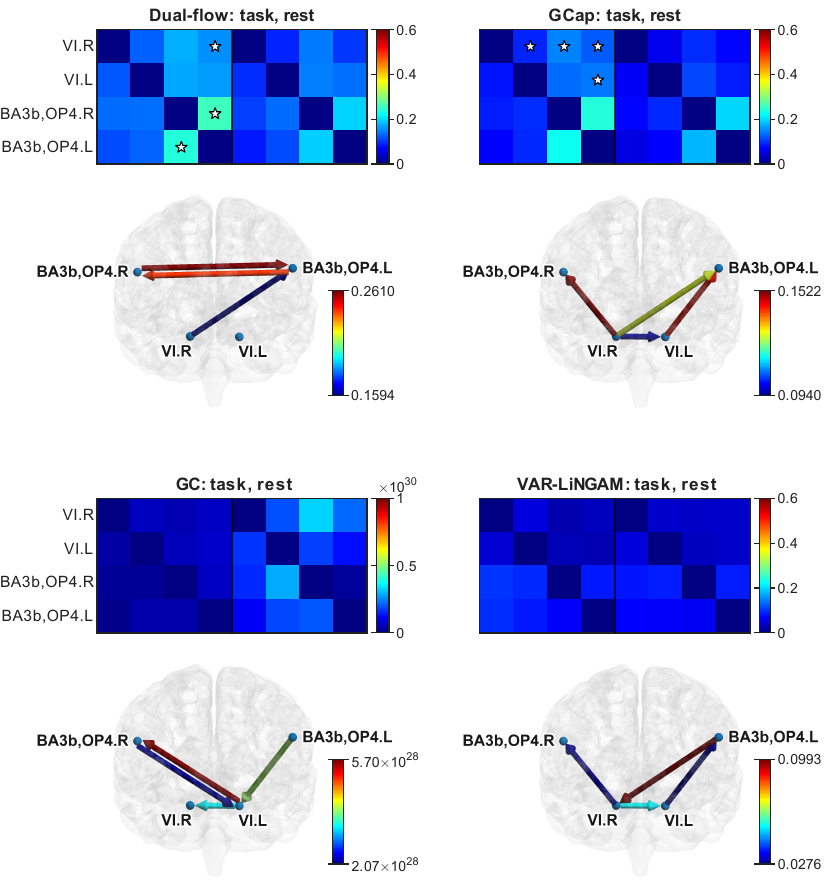}  
 \caption{\textbf{Connectivity matrices of Dual-flow (upper left), GCap (upper right), GC (lower left), and VAR-LiNGAM (lower right) during tongue motion and rest.} Significant task-related directed interactions are marked with stars and visualized in BrainNet Viewer \citep{xia2013brainnet}. Since no edges met the task-sensitivity criteria for GC or VAR-LiNGAM, the four nonsignificant edges with the lowest FDR-adjusted $p$-values are visualized for each method. GIMME was excluded because the task segments were shorter than its documented minimum timeseries length.}
    \label{fig:brainnet_all}
\end{figure}

To quantify task sensitivity, we compared tongue-task and rest EC estimates for each directed ROI pair across the 44 participants, treating participants as the unit of inference. Using the same criterion across estimators, we defined task-sensitive directed interactions as ROI pairs with low between-subject variability (CV below the 30th percentile) and a significant task-related increase relative to rest, assessed using right-tailed paired tests followed by false-discovery-rate correction ($q<0.1$). Per-pair results for all four estimators are reported in Table~\ref{tab:tongue_stats}. Three directed edges met these criteria for Dual-flow and four for GCap; none met the criteria for GC or VAR-LiNGAM. Fig.~\ref{fig:brainnet_all} displays the identified Dual-flow and GCap edges on the brain, alongside the four nonsignificant edges with the lowest FDR-adjusted $p$-values for each of GC and VAR-LiNGAM. These 16-point task segments provide a practical stress test under limited observations, in which Dual-flow detected task-sensitive directed interactions consistent with the literature-supported circuitry described in Section~\ref{sec:tongue-motion}.

\begin{table}[h!]
    \centering
    \caption{Student's $t$-test positive false discovery rate (pFDR) and coefficient of variation (CV) comparing Dual-flow, GCap, GC, and VAR-LiNGAM between task and rest conditions for each directed ROI pair in the HCP tongue task dataset. A directed connection is considered task-related if pFDR $<0.1$ and CV $<$ 30\% percentile; values meeting each threshold are shown in bold.}
    \label{tab:tongue_stats}
    \resizebox{\textwidth}{!}{
    \begin{tabular}{llcccccccc}
    \toprule
    \multirow{2}{*}{Sender ROI} & \multirow{2}{*}{Receiver ROI}
     & \multicolumn{2}{c}{Dual-flow} & \multicolumn{2}{c}{GCap} & \multicolumn{2}{c}{GC} & \multicolumn{2}{c}{VAR-LiNGAM} \\
    \cmidrule(lr){3-4} \cmidrule(lr){5-6} \cmidrule(lr){7-8} \cmidrule(lr){9-10}
     & & pFDR & CV & pFDR & CV & pFDR & CV & pFDR & CV \\
    \midrule
    VI.R       & VI.L       & .1101          & \textbf{3.6144} & \textbf{.0915} & \textbf{3.7060} & .9966 & 3.9374          & .2715 & \textbf{3.7033} \\
    VI.R       & BA3b,OP4.R & .1413          & 4.1552          & \textbf{.0767} & \textbf{3.1041} & .9966 & 4.5693          & .9912 & \textbf{2.6882} \\
    VI.R       & BA3b,OP4.L & \textbf{.0959} & \textbf{2.6447} & \textbf{.0438} & \textbf{2.6062} & .9966 & 4.0495          & .9880 & 8.2120 \\
    VI.L       & VI.R       & .1537          & 5.1530          & .1963          & 7.1669          & .9966 & \textbf{3.5852} & .9880 & 10.9397 \\
    VI.L       & BA3b,OP4.R & .1570          & 5.8323          & .1963          & 6.7362          & .9966 & \textbf{2.9687} & .9880 & 7.0685 \\
    VI.L       & BA3b,OP4.L & .1537          & 4.9825          & \textbf{.0438} & \textbf{2.5106} & .9966 & 6.4292          & .9912 & \textbf{2.7151} \\
    BA3b,OP4.R & VI.R       & .1570          & 5.6009          & .1963          & 6.4526          & .9966 & 4.0842          & .3050 & 5.1349 \\
    BA3b,OP4.R & VI.L       & .5562          & 46.6850         & .4677          & 81.3787         & .9966 & \textbf{2.3304} & .6447 & 11.2962 \\
    BA3b,OP4.R & BA3b,OP4.L & \textbf{.0985} & \textbf{3.2781} & .0915 & 4.0262          & .9966 & 5.8199          & .9779 & 37.1344 \\
    BA3b,OP4.L & VI.R       & .1537          & 4.9899          & .1961          & 5.8293          & .9966 & 4.1199          & .2715 & \textbf{3.9087} \\
    BA3b,OP4.L & VI.L       & .3076          & 11.4084         & {.0915} & 4.0120          & .9966 & \textbf{3.6104} & .2715 & 4.3630 \\
    BA3b,OP4.L & BA3b,OP4.R & \textbf{.0985} & \textbf{3.2164} & {.0915} & 3.8329          & .9966 & 4.1509          & .6447 & 14.2828 \\
    \bottomrule
    \end{tabular}}
\end{table}

This application was designed to test two related questions. First, does Dual-flow detect stronger directed interactions during tongue movement than during rest in physiologically plausible motor circuits? Second, compared with GCap, GC, and VAR-LiNGAM, does Dual-flow provide greater sensitivity when the fitted residuals for tongue-motion trials show stronger departure from Gaussianity? In the main text, we therefore focus on whether the tongue condition---which showed the strongest departure from Gaussianity among the movement conditions---also exhibits the clearest practical benefit from distribution-aware capacity estimation.

\subsection{Cross-Modal Correspondence of Time-Varying EC Estimates}
\label{app:swc}
To assess the physiological relevance of time-varying EC estimates, we analyzed the concurrent rat BOLD--LFP recordings and six LFP band-limited power (BLP) time series described in Appendix~\ref{app:gaussianity_analysis}. We used 50-s sliding windows, following \citet{thompson2013neural}, with 50\% overlap between consecutive windows (25-s steps). For a fair comparison, all estimators were fitted independently within each window using the same window length and overlap. For each estimator, we estimated directed connectivity from left to right S1 in both modalities, then pooled the windowed estimates across scans and correlated the BLP-derived and corresponding BOLD-derived connectivity sequences. We assessed cross-modal correspondence using 10{,}000 bootstrap resamples of the 22 scans, retaining all paired windowed estimates within each resampled scan. Statistical significance was assessed using one-sided bootstrap tests of positive correlation, with Benjamini--Hochberg correction across the six bands within each method.

\begin{figure}[h!]
    \centering
    \includegraphics[width=300pt]{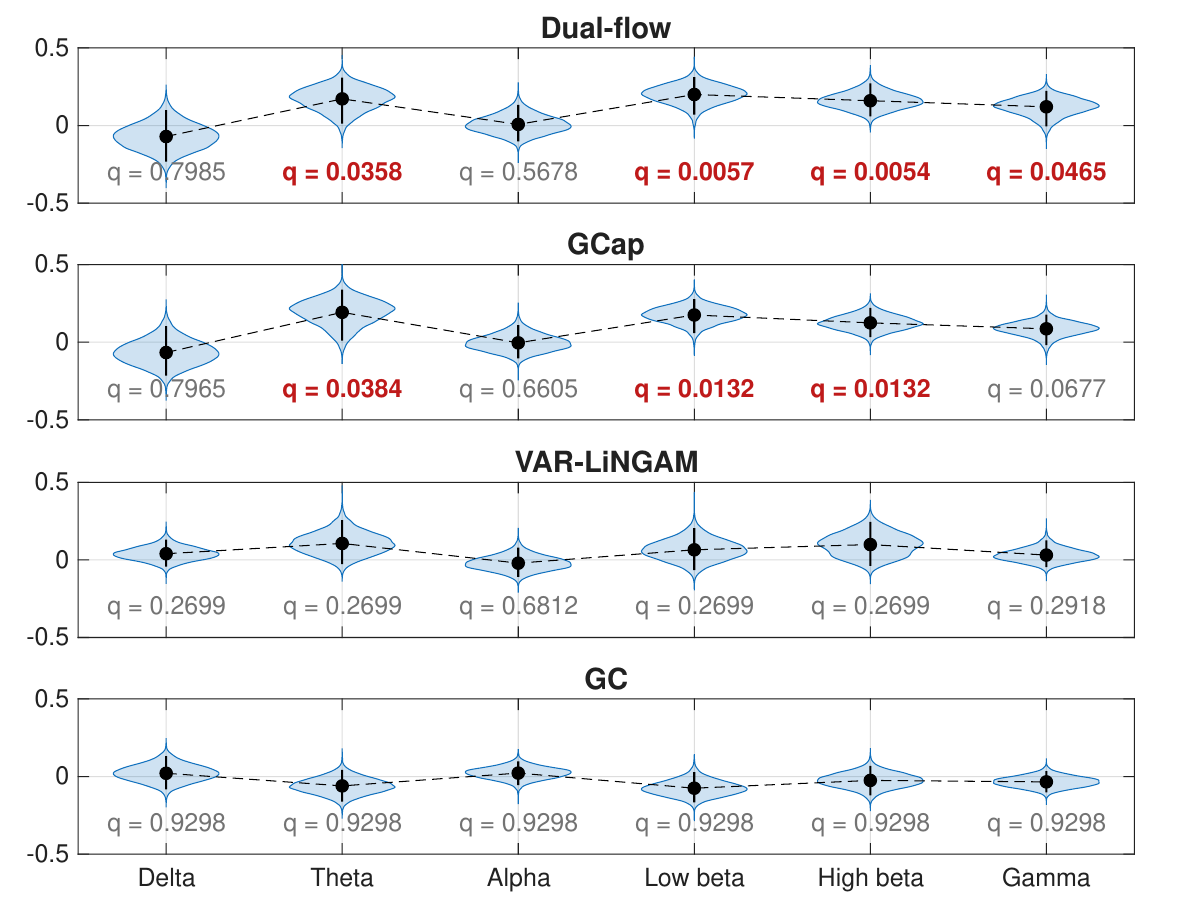}
    \caption{\textbf{Cross-modal correspondence of time-resolved EC estimates across six LFP frequency bands.} Each panel shows one estimator. Violins represent correlations between BLP- and BOLD-derived EC estimates (left S1$\to$right S1), pooled across windows, over 10{,}000 bootstrap resamples of the 22 scans. Black markers and vertical bars indicate bootstrap means and 95\% percentile intervals. Values below the violins are one-sided bootstrap $q$ values after Benjamini--Hochberg correction across bands within each method; significant values ($q<0.05$) appear in bold red.}
    \label{fig:swc}
\end{figure}

The band-specific correspondence between Dual-flow-derived BOLD and BLP EC estimates matches that previously reported using sliding-window correlation \citep[Fig.~2A]{thompson2013neural}, with significant correspondence in theta, both beta bands, and gamma. GCap showed a similar pattern but did not reach significance in gamma ($q=0.068$), whereas GC and VAR-LiNGAM showed no significant correspondence in any band (Fig.~\ref{fig:swc}). This analysis provides cross-modal support, beyond simulation, that time-resolved capacity-based EC estimates capture shared temporal variation in directed interactions across hemodynamic and electrophysiological measurements.

\section{Ground-Truth Effective-Connectivity Validation in Brain-Like Simulations}
\label{app:simulation_details}

\subsection{Simulation Design}
\label{app:sim_topologies}
We conducted controlled stress tests using a DCM--Balloon simulation framework adapted from \citet{smith2011network}, with prespecified ground-truth directed EC matrices. Linear neural dynamics were driven by two-state Markov-switching Gaussian inputs, with off/on means of 0/1, a common SD of 0.05, and off-to-on/on-to-off transition rates of 0.1/0.4~s$^{-1}$. A nonlinear Balloon--Windkessel model transformed neural activity into BOLD signals. We generated 5-min recordings with \(TR=1\) s and added temporally correlated measurement noise using an AR(1) process (\(\rho=0.4\)) with two-component Gaussian-mixture innovations. 
The resulting noise was scaled to an overall SNR of 10 dB across observed ROIs and realizations.

We evaluated ten conditions spanning common drivers, indirect pathways, hidden nodes, feedback, modular organization, heterogeneous hemodynamics, and increasing network size. Topologies were drawn or adapted from prior benchmarks \citep{xu2017initial,smith2011network,sanchezromero2019estimating} or constructed for this study; their sources, directed edges, coupling weights, and modifications are specified in Table~\ref{tab:sim_topologies}. We generated 50 realizations per condition, keeping the ground-truth connectivity matrix fixed while varying neural inputs and measurement noise. Hemodynamic parameters additionally varied across realizations in the conditions marked with an asterisk.

In the heterogeneous-HRF conditions, regional BOLD delay offsets received Gaussian perturbations with SD \(=0.10\) s, and regional transit times received multiplicative lognormal perturbations with approximately \(5\%\) CV before clipping. Delay offsets were bounded to \(\pm1.50\) s, and transit times to 0.65--1.40 s. These manipulations were motivated by reported between-region and between-subject HRF heterogeneity \citep{handwerker2004variation,smith2011network}.
\begin{table}[h!]
\centering
\caption{Simulation topologies and manipulations. ROIs are given as observed/total, so that conditions with a hidden node have fewer observed than total ROIs. Asterisks mark conditions in which hemodynamic response functions varied across realizations.}
\label{tab:sim_topologies}
\resizebox{\textwidth}{!}{
\begin{tabular}{lccp{4.2cm}p{4.0cm}p{3.4cm}}
\toprule
Condition & ROIs & Edges & Ground-truth topology & Manipulation tested & Topology source \\
\midrule
CD-e & 3/3 & 2 &
$1\!\to\!2$, $1\!\to\!3$; weights 0.4/0.4 &
Equal-coupling common driver &
\citet{xu2017initial} \\
\addlinespace
CD-u & 3/3 & 2 &
$1\!\to\!2$, $1\!\to\!3$; weights 0.8/0.4 &
Unequal-coupling common driver &
Adapted from \citet{xu2017initial} \\
\addlinespace
Diamond & 5/5 & 5 &
$1\!\to\!2$, $1\!\to\!3$, $2\!\to\!4$, $3\!\to\!4$, $4\!\to\!5$; weights 0.4 &
Parallel indirect paths and convergence &
Present study \\
\addlinespace
Diamond (H) & 4/5 & 5 total; 3 evaluated &
Same diamond; ROI 1 simulated but omitted from the analysis &
Hidden common driver of ROIs 2 and 3 &
Present-study extension \\
\addlinespace
Chain & 5/5 & 5 &
$1\!\to\!2\!\to\!3\!\to\!4\!\to\!5$ and $1\!\to\!5$; weights 0.4 &
Indirect path with direct shortcut &
Smith S5 \citep{smith2011network} \\
\addlinespace
Chain (HRF)* & 5/5 & 5 &
Same as Chain &
Regional HRF delays $[-0.60, 0.70]$ s; transit multipliers $[0.85, 1.18]$ &
Smith S5 with HRF manipulation \\
\addlinespace
Feedback & 5/5 & 6 &
$1\!\to\!2\!\to\!3\!\to\!4\!\to\!5$, $3\!\to\!2$, $5\!\to\!3$; weights 0.4 &
Reciprocal loop $2\!\leftrightarrow\!3$ and cycle $3\!\to\!4\!\to\!5\!\to\!3$ &
Present-study recurrent extension \\
\addlinespace
Mod10 & 10/10 & 11 &
Two S5 subnetworks connected by $3\!\to\!8$ &
Two-module organization &
Smith S10 \citep{smith2011network} \\
\addlinespace
Comp12* & 10/12 & 22 total; 18 evaluated &
Module A: $1\!\to\!2\!\to\!3\!\to\!4\!\to\!5$, $3\!\to\!2$, $5\!\to\!3$, $1\!\to\!4$, $2\!\to\!5$. Module B: analogous over ROIs 6--10. Cross-module: $5\!\to\!6$, $9\!\to\!2$. Hidden drivers: $11\!\to\!\{2,5\}$, $12\!\to\!\{7,10\}$; weights 0.4 &
Indirect paths, feedback, modular structure, two hidden drivers, and heterogeneous HRFs combined &
Present-study combined topology \\
\addlinespace
Macq28* & 28/28 & 52 &
Fixed weighted SmallDegree matrix; weights 0.468--0.551; 10 cycles including five reciprocal pairs &
Larger recurrent network with independently varying regional HRFs &
SmallDegree topology \citep{sanchezromero2019estimating}, pruned from a 28-node subnetwork of the macaque cortical connectome \citep{markov2014weighted} \\
\bottomrule
\end{tabular}}
\end{table}

\subsection{Evaluation Procedure and Performance}
\label{app:sim_protocol}
\label{app:sim_full}

\paragraph{Evaluation Procedure.}
All methods were evaluated on the same 50 simulated datasets per condition against the same prespecified directed-edge ground truth. We evaluated all \(N(N-1)\) ordered edges independently, allowing neither, one, or both directions within each ROI pair to be present. For each realization, AUROC and AUPRC were computed from all ordered-edge scores. Precision, sensitivity, and false-positive rate (FPR) were computed using a method-specific threshold selected from the other 49 realizations within the same condition by maximizing the Matthews correlation coefficient (leave-one-simulation-out, LOSO-MCC). The same calibration procedure was applied to every method-condition pair, excluding the evaluated realization from threshold selection.

\begin{table}[h!]
\centering
\caption{Directed-edge recovery in brain-like simulations. Entries are mean \(\pm\) SD across 50 realizations. Precision, sensitivity, and FPR use LOSO-MCC-calibrated thresholds. Bold marks the best mean in each metric within a condition (highest for AUROC, AUPRC, precision, and sensitivity; lowest for FPR). GIMME was not evaluated on Macq28 because its 28 ROIs exceed the recommended range of 3--20 ROIs \citep{beltz2017network}.}
\label{tab:sim_full}
\scriptsize
\begin{tabular*}{\textwidth}{@{\extracolsep{\fill}}llccccc@{}}
\toprule
Condition & Method & AUROC & AUPRC & Precision & Sensitivity & FPR \\
\midrule
\multirow{5}{*}{CD-e}
 & Dual-flow  & \textbf{.888$\pm$.108} & \textbf{.857$\pm$.132} & \textbf{.887$\pm$.206} & .620$\pm$.216 & .070$\pm$.134 \\
 & GCap       & .700$\pm$.129 & .623$\pm$.155 & .472$\pm$.069 & \textbf{.960$\pm$.137} & .545$\pm$.120 \\
 & VAR-LiNGAM & .725$\pm$.315 & .739$\pm$.273 & .618$\pm$.404 & .570$\pm$.378 & .200$\pm$.226 \\
 & GIMME      & .498$\pm$.018 & .333$\pm$.000 & .000$\pm$.000 & .000$\pm$.000 & \textbf{.005$\pm$.035} \\
 & GC         & .580$\pm$.221 & .572$\pm$.201 & .332$\pm$.206 & .550$\pm$.323 & .555$\pm$.249 \\
\midrule
\multirow{5}{*}{CD-u}
 & Dual-flow  & \textbf{.858$\pm$.104} & \textbf{.830$\pm$.112} & .890$\pm$.273 & .470$\pm$.120 & .040$\pm$.093 \\
 & GCap       & .668$\pm$.130 & .579$\pm$.125 & .455$\pm$.089 & \textbf{.860$\pm$.227} & .525$\pm$.154 \\
 & VAR-LiNGAM & .713$\pm$.318 & .726$\pm$.270 & .568$\pm$.400 & .570$\pm$.404 & .220$\pm$.206 \\
 & GIMME      & .750$\pm$.000 & .667$\pm$.000 & \textbf{1.000$\pm$.000} & .500$\pm$.000 & \textbf{.000$\pm$.000} \\
 & GC         & .610$\pm$.187 & .576$\pm$.166 & .395$\pm$.128 & .810$\pm$.265 & .630$\pm$.184 \\
\midrule
\multirow{5}{*}{Diamond}
 & Dual-flow  & \textbf{.874$\pm$.076} & \textbf{.785$\pm$.120} & \textbf{.853$\pm$.168} & .532$\pm$.234 & .040$\pm$.049 \\
 & GCap       & .777$\pm$.050 & .541$\pm$.065 & .418$\pm$.072 & \textbf{.820$\pm$.168} & .404$\pm$.154 \\
 & VAR-LiNGAM & .751$\pm$.126 & .618$\pm$.186 & .605$\pm$.217 & .560$\pm$.236 & .131$\pm$.089 \\
 & GIMME      & .505$\pm$.032 & .259$\pm$.047 & .040$\pm$.198 & .012$\pm$.063 & \textbf{.003$\pm$.013} \\
 & GC         & .607$\pm$.157 & .453$\pm$.165 & .374$\pm$.173 & .508$\pm$.195 & .312$\pm$.144 \\
\midrule
\multirow{5}{*}{Diamond (H)}
 & Dual-flow  & \textbf{.910$\pm$.079} & \textbf{.844$\pm$.120} & \textbf{.927$\pm$.154} & .560$\pm$.218 & .022$\pm$.045 \\
 & GCap       & .799$\pm$.070 & .591$\pm$.103 & .462$\pm$.066 & \textbf{.847$\pm$.204} & .344$\pm$.135 \\
 & VAR-LiNGAM & .787$\pm$.172 & .675$\pm$.220 & .620$\pm$.247 & .660$\pm$.247 & .149$\pm$.111 \\
 & GIMME      & .502$\pm$.037 & .260$\pm$.049 & .000$\pm$.000 & .000$\pm$.000 & \textbf{.009$\pm$.030} \\
 & GC         & .589$\pm$.233 & .494$\pm$.226 & .394$\pm$.279 & .507$\pm$.303 & .311$\pm$.200 \\
\midrule
\multirow{5}{*}{Chain}
 & Dual-flow  & \textbf{.938$\pm$.042} & \textbf{.876$\pm$.069} & \textbf{.764$\pm$.166} & .688$\pm$.262 & .103$\pm$.098 \\
 & GCap       & .761$\pm$.056 & .504$\pm$.065 & .430$\pm$.076 & \textbf{.744$\pm$.218} & .355$\pm$.165 \\
 & VAR-LiNGAM & .715$\pm$.160 & .603$\pm$.200 & .606$\pm$.326 & .420$\pm$.253 & .101$\pm$.096 \\
 & GIMME      & .522$\pm$.051 & .287$\pm$.071 & .230$\pm$.419 & .052$\pm$.097 & \textbf{.008$\pm$.022} \\
 & GC         & .603$\pm$.100 & .428$\pm$.118 & .347$\pm$.183 & .384$\pm$.217 & .260$\pm$.141 \\
\midrule
\multirow{5}{*}{Chain (HRF)}
 & Dual-flow  & \textbf{.939$\pm$.057} & \textbf{.891$\pm$.091} & .783$\pm$.184 & .776$\pm$.216 & .101$\pm$.122 \\
 & GCap       & .795$\pm$.059 & .615$\pm$.086 & .395$\pm$.068 & \textbf{.876$\pm$.127} & .472$\pm$.154 \\
 & VAR-LiNGAM & .693$\pm$.134 & .511$\pm$.144 & .354$\pm$.137 & .568$\pm$.266 & .337$\pm$.155 \\
 & GIMME      & .605$\pm$.024 & .409$\pm$.036 & \textbf{.990$\pm$.071} & .208$\pm$.040 & \textbf{.001$\pm$.009} \\
 & GC         & .752$\pm$.099 & .631$\pm$.118 & .804$\pm$.320 & .208$\pm$.090 & .032$\pm$.062 \\
\midrule
\multirow{5}{*}{Feedback}
 & Dual-flow  & \textbf{.889$\pm$.071} & \textbf{.866$\pm$.091} & \textbf{.769$\pm$.160} & \textbf{.790$\pm$.183} & .202$\pm$.162 \\
 & GCap       & .613$\pm$.043 & .534$\pm$.051 & .476$\pm$.040 & .748$\pm$.144 & .555$\pm$.134 \\
 & VAR-LiNGAM & .563$\pm$.147 & .545$\pm$.141 & .511$\pm$.243 & .305$\pm$.173 & .192$\pm$.108 \\
 & GIMME      & .333$\pm$.006 & .400$\pm$.000 & .000$\pm$.000 & .000$\pm$.000 & \textbf{.000$\pm$.000} \\
 & GC         & .514$\pm$.111 & .495$\pm$.104 & .411$\pm$.058 & .750$\pm$.182 & .715$\pm$.165 \\
\midrule
\multirow{5}{*}{Mod10}
 & Dual-flow  & \textbf{.949$\pm$.027} & \textbf{.815$\pm$.073} & \textbf{.802$\pm$.151} & .616$\pm$.205 & .029$\pm$.030 \\
 & GCap       & .883$\pm$.026 & .469$\pm$.055 & .390$\pm$.054 & \textbf{.782$\pm$.149} & .178$\pm$.063 \\
 & VAR-LiNGAM & .701$\pm$.098 & .393$\pm$.146 & .422$\pm$.170 & .396$\pm$.148 & .084$\pm$.046 \\
 & GIMME      & .517$\pm$.034 & .153$\pm$.055 & .230$\pm$.384 & .038$\pm$.064 & \textbf{.005$\pm$.010} \\
 & GC         & .630$\pm$.098 & .266$\pm$.090 & .219$\pm$.095 & .411$\pm$.173 & .218$\pm$.083 \\
\midrule
\multirow{5}{*}{Comp12}
 & Dual-flow  & \textbf{.879$\pm$.025} & \textbf{.724$\pm$.052} & .708$\pm$.113 & .581$\pm$.083 & .066$\pm$.036 \\
 & GCap       & .852$\pm$.023 & .644$\pm$.039 & .542$\pm$.063 & \textbf{.656$\pm$.077} & .143$\pm$.041 \\
 & VAR-LiNGAM & .692$\pm$.062 & .428$\pm$.080 & .409$\pm$.099 & .426$\pm$.124 & .160$\pm$.063 \\
 & GIMME      & .566$\pm$.005 & .313$\pm$.019 & \textbf{.870$\pm$.188} & .088$\pm$.028 & \textbf{.005$\pm$.007} \\
 & GC         & .736$\pm$.044 & .483$\pm$.060 & .446$\pm$.093 & .441$\pm$.158 & .157$\pm$.090 \\
\midrule
\multirow{4}{*}{Macq28}
 & Dual-flow  & \textbf{.884$\pm$.018} & \textbf{.521$\pm$.042} & \textbf{.573$\pm$.118} & .453$\pm$.105 & \textbf{.029$\pm$.018} \\
 & GCap       & .846$\pm$.021 & .391$\pm$.037 & .427$\pm$.082 & .403$\pm$.074 & .043$\pm$.019 \\
 & VAR-LiNGAM & .578$\pm$.038 & .110$\pm$.022 & .102$\pm$.020 & .305$\pm$.078 & .200$\pm$.046 \\
 & GC         & .679$\pm$.029 & .145$\pm$.022 & .122$\pm$.020 & \textbf{.568$\pm$.125} & .315$\pm$.102 \\
\bottomrule
\end{tabular*}
\end{table}

\paragraph{Performance and reliability.} Table~\ref{tab:sim_full} reports the mean and standard deviation of all five metrics across the 50 realizations per condition. Dual-flow achieved the highest mean AUROC and AUPRC, the two threshold-independent ranking metrics, among the evaluated methods across all ten conditions (AUROC: .858--.949; AUPRC: .521--.891), including Comp12 (.879/.724) and Macq28 (.884/.521). Macq28 contains only 52 true edges among 756 candidates, yielding a chance-level AUPRC reference of .069. Across 50 realizations, AUROC/AUPRC were \(.888\pm.108/.857\pm.132\) for CD-e and \(.858\pm.104/.830\pm.112\) for CD-u. Threshold-dependent metrics---precision, sensitivity, and FPR---often showed greater variability, particularly in these common-driver conditions. Each realization contains only two true and four null edges, so sensitivity and FPR change in increments of .50 and .25, respectively. LOSO-MCC calibration yielded precision/FPR of .887/.070 for CD-e and .890/.040 for CD-u, with sensitivity of .620 and .470, respectively. This coarse metric resolution partly explains the larger threshold-dependent SDs, while mean edge-ranking performance remained strong. GCap achieved higher sensitivity in several conditions but also substantially higher FPR, whereas GIMME generally showed low FPR alongside low sensitivity. Overall, Dual-flow maintained strong directed-edge recovery across individual stressors, their coexistence, and the larger network.

\section{Computational Requirements}
\label{app:cost}
Dual-flow performs iterative optimization of the generator and observer flows, while independent directed-pair evaluations can be executed concurrently. As a representative benchmark, we evaluated the Mod10 workload comprising 900 directed EC evaluations (10 realizations $\times$ 90 directed pairs) using 4 $\times$ NVIDIA L40S GPUs on an AMD EPYC 7763 64-core system with 1.0 TiB RAM, with eight pair evaluations scheduled concurrently per GPU. Across five seeded repetitions, the complete workload required $1272.74\pm122.57$ s ($\sim$21.2 min) of wall time, corresponding to an amortized $1.41\pm0.14$ s per directed EC evaluation. We used $L=256$, for which capacity estimates were stable across the evaluated training-sequence lengths (Appendix~\ref{app:L_sensitivity}). The current implementation is well suited to hypothesis-driven analyses of prespecified circuits and moderate-sized subnetworks, an established practice in effective-connectivity studies across cognitive and clinical neuroimaging \citep{smith2012effective,deshpande2012investigating,bielczyk2019disentangling}.

\end{document}